\documentclass[preprint,3p,authoryear]{elsarticle}

\usepackage{array}      
\usepackage{ragged2e}
\usepackage{multirow}
\usepackage{tikz}
\usepackage{amssymb}
\usepackage{subcaption}
\usepackage{amsmath}
\usepackage{url}
\usepackage{color}
\definecolor{darkgreen}{rgb}{0.0, 0.5, 0.0}
\definecolor{lightblue}{rgb}{0.68, 0.85, 0.9}
\usepackage{algorithm} 
\usepackage{algorithmic} 
\usepackage{amsmath,amssymb,amsfonts} 
\usepackage{mathrsfs} 
\usepackage{graphicx}
\usepackage{lineno}
\usepackage{textcomp}
\usepackage{relsize} 
\usepackage{cuted}
\usepackage{dblfloatfix} 
\usepackage[bookmarks,bookmarksnumbered]{hyperref} 
\hypersetup{colorlinks = true,
    linkcolor=blue,
    filecolor=magenta,      
    urlcolor=cyan}
\usepackage[labelfont=bf]{caption} 
\usepackage[nameinlink,capitalize]{cleveref} 
\usepackage{flushend} 
\usepackage{booktabs} 
\usepackage[bb=boondox]{mathalfa} 
\usepackage{float} 
\usepackage{makecell} 
\usepackage{setspace}
\usepackage[english]{babel}
\usepackage[authoryear]{natbib}

\definecolor{inchworm}{rgb}{0.7, 0.93, 0.36} 
\definecolor{orng}{rgb}{1.0, 0.31, 0.0} 

\newcommand\eatpunct[1]{} 
\addto\extrasenglish{

} 
\addto\captionsenglish{} 

\journal{Elsevier}

\begin{document}

\begin{titlepage}
\begin{center}
\vspace*{10pt}
\doublespacing
{\Large A Multi-Temporal Multi-Spectral Attention-Augmented Deep Convolution Neural Network with Contrastive Learning for Crop Yield Prediction}
\vspace*{25pt}

Shalini Dangi$^{a}$ (phd2301201003@iiti.ac.in), Surya Karthikeya Mullapudi$^{a}$ (cse210001041@iiti.ac.in), Chandravardhan Singh Raghaw$^{a}$ (phd2201101016@iiti.ac.in), Shahid Shafi Dar$^{a}$ (phd2201201004@iiti.ac.in), Mohammad Zia Ur Rehman$^{a}$ (phd2101201005@iiti.ac.in), Nagendra Kumar$^{a}$ (nagendra@iiti.ac.in) \\

\hspace{10pt}

\begin{flushleft}
\small  
$^a$Department of Computer Science and Engineering, Indian Institute of Technology Indore, Indore 453552, India \\
\vspace{10mm}
This is the preprint of the accepted paper.\\
The paper is published in \textbf{Computers and Electronics in Agriculture (COMPAG).}\\
The published version is available at: \url{https://doi.org/10.1016/j.compag.2025.110895}


\end{flushleft}        
\end{center}
\end{titlepage}


\begin{frontmatter}

\title{A Multi-Temporal Multi-Spectral Attention-Augmented Deep Convolution Neural Network with Contrastive Learning for Crop Yield Prediction}
\author{Shalini Dangi}
\ead{phd2301201003@iiti.ac.in}

\author{Surya Karthikeya Mullapudi}
\ead{cse210001041@iiti.ac.in}

\author{Chandravardhan Singh Raghaw}
\ead{phd2201101016@iiti.ac.in}

\author{Shahid Shafi Dar}
\ead{phd2201201004@iiti.ac.in}

\author{Mohammad Zia Ur Rehman}
\ead{phd2101201005@iiti.ac.in}

\author{Nagendra Kumar \corref{cor1}}
\cortext[cor1]{Corresponding author email: \textit{nagendra@iiti.ac.in}}

\address{Department of Computer Science and Engineering, Indian Institute of Technology (IIT) Indore, Indore 453552, India}

\begin{abstract}
\doublespacing
Precise yield prediction is essential for agricultural sustainability and food security. However, climate change complicates accurate yield prediction by affecting major factors such as weather conditions, soil fertility, and farm management systems. Advances in technology have played an essential role in overcoming these challenges by leveraging satellite monitoring and data analysis for precise yield estimation. Current methods rely on spatio-temporal data for predicting crop yield, but they often struggle with multi-spectral data, which is crucial for evaluating crop health and growth patterns. To resolve this challenge, we propose a novel \textbf{M}ulti-\textbf{T}emporal \textbf{M}ulti-\textbf{S}pectral \textbf{Yield} prediction \textbf{Net}work, MTMS-YieldNet, that integrates spectral data with spatio-temporal information to effectively capture the correlations and dependencies between them. While existing methods that rely on pre-trained models trained on general visual data, MTMS-YieldNet utilizes contrastive learning for feature discrimination during pre-training, focusing on capturing spatial-spectral patterns and spatio-temporal dependencies from remote sensing data. Contrastive learning finds the relative features that distinguish crop growth patterns over different temporal intervals. The stacked attention mechanism is applied to improve spectral-spatial feature extraction by focusing on the most important spectral bands and spatial regions, further enhancing forecasting precision. We use an optimization approach inspired by natural balance processes to identify key spectral and temporal features for effective feature selection in crop yield prediction. We evaluate MTMS-YieldNet on various datasets using remote sensing images from Sentinel-1, Sentinel-2, and Landsat-8, treating each source as a distinct dataset to capture diverse agricultural patterns. Both quantitative and qualitative assessments highlight the excellence of the proposed MTMS-YieldNet over seven existing state-of-the-art methods.  For the Sentinel-1 dataset, the MTMS-YieldNet achieves 0.336 MAPE, 0.497 RMSLE, and 0.362 SMAPE, and for the Landsat-8 dataset, it achieves 0.353 MAPE, 0.511 RMSLE, and 0.428 SMAPE.  On Sentinel-2, it achieves an outstanding performance of 0.331 MAPE, 0.589 RMSLE, and 0.433 SMAPE, demonstrating its effectiveness in yield prediction across varying climatic and seasonal conditions in this agriculturally significant region.  The outstanding performance of MTMS-YieldNet improves yield predictions and provides valuable insights that can assist farmers in making better decisions, potentially improving crop yields.

\end{abstract}

\end{frontmatter}

\section{Introduction}
\label{introduction}
Agriculture is a crucial sector worldwide, significantly impacting the economies of many countries. Agriculture holds considerable importance in India, 
representing around 18\% of the country's GDP and providing employment to around 50\% of the workforce,
as reported in the Economic Survey of India 2024\footnote{ 
\url{https://www.indiabudget.gov.in/economicsurvey/}}.
 The exponential growth of the population is motivated to enhance the productivity of agriculture owing to the rising demand for food, the challenges in producing it, and the adverse impacts on the environment. 
 By 2050, the Earth's population is about to reach approximately 9.6 billion, implying that food production must increase by 70\%. 
 Additionally, climate change also emerged as a challenging problem in agriculture, affecting soil health, seasonal variations, and plant growth \citep{Zhao_2017}. The combined effects of population growth and atmospheric change may pose a threat to global food assurance. There is a need for climate change-aware crop yield prediction, which optimizes agricultural production and also ensures adequate food supply \citep{Lin_2023}.
 
Conventional methods in yield prediction depend on manual crop sample collection that covers a small portion of the cropland. These methods often require manual efforts which are costly that leads to inaccuracies. Over a couple of decades, significant research work has explored the integration of terrestrial and remote-monitoring approaches for automating and improving yield prediction for different crops. 
These advanced sensing approaches improve spatial and temporal resolution and enhance prediction performance by precise crop management across the growing phase, which provides a more dynamic and accurate understanding of crop health and yield potential. The remote sensing approach is most important in yield forecasting because of its ability to cover large areas to monitor crop dynamics over time and also provide precise performance in comparison to ground-based approaches \citep{Jung_2021}. 
The spatio-temporal scaling effect also plays an important role in yield estimation. The understanding of high-resolution images is essential for effective crop growth monitoring \citep{Yuan_2020}. Various methods have been applied in crop yield prediction with the advancement of technology including multiple linear regression \citep{Piekutowska_2021}, and support vector regression \citep{Zhang_2020}. Machine learning methods face difficulties in recognizing long-term patterns and managing inconsistencies in data from multiple sources. 

The combination of remote-sensing data and advanced deep learning approaches provide enhancement in prediction with agricultural variations. \cite{Fan_2022} proposed a GNN-RNN method for predicting crop yields in which temporal data integrates with geospatial data by utilizing RNNs for temporal integration and GNNs to account for geographical relationships.  Also attention based \citep{Qu_2024} and contrastive learning based \citep{Seong_2024} mechanism introduced for identifying the relevant region and correlation between features. 
These approaches effectively identified patterns between spatio-temporal features but failed to capture complex patterns with spectral information. The availability of limited spectral information became a hurdle in precise yield prediction and failed to detect variations in crop health and growth that may lead to inaccurate forecasts. There is need to improve the spectral resolution for precise monitoring but it is still challenging.
\begin{figure}[h!]
    \centering
    \captionsetup{justification=centering}
    \includegraphics[width=0.75\textwidth]{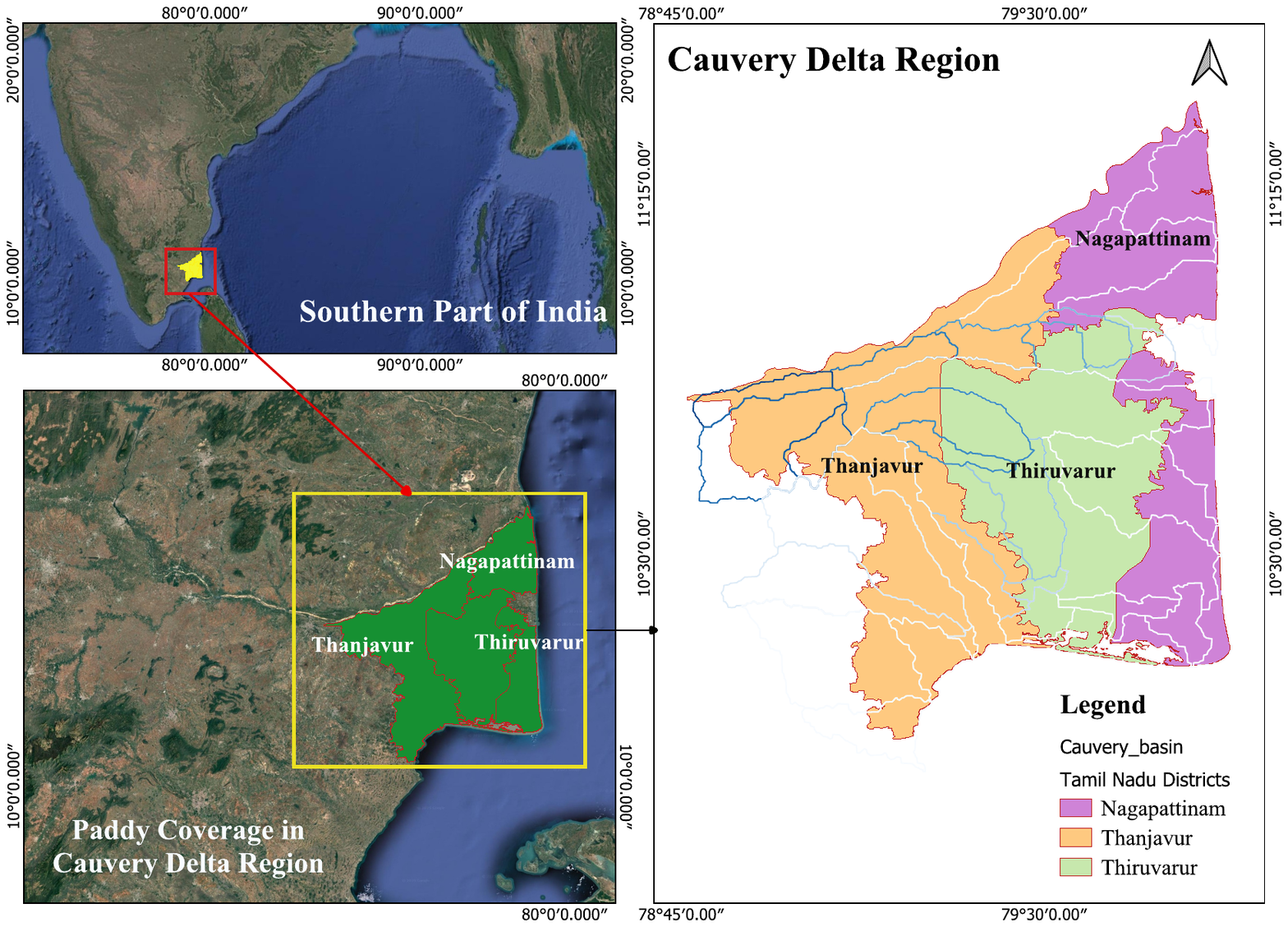}
    \caption{Cauvery Delta: Study region for paddy crop analysis using multi-spectral and multi-temporal data.}
    \label{fig: Cauvery}
\end{figure}
In our research, we utilize satellite images covering the Cauvery Delta region known as the Rice Bowl of Tamil Nadu, as illustrated in Fig.~\ref{fig: Cauvery}. The Cauvery Delta is famous for its fertile alluvial soil and extensive rice fields. The crop fields for our work belong to the following districts:
Thanjavur, Thiruvarur, and Nagapattinam.

To enhance the performance of yield prediction and mitigate the challenges identified in existing studies, this paper introduces the Multi-Temporal Multi-Spectral Yield Prediction Network (MTMS-YieldNet), a novel framework designed to predict crop yield. MTMS-YieldNet employs a data-driven deep-learning approach, incorporating novel techniques such as Spectral-Spatial Attention Module, Spatio-Temporal Dependency Module,  Spatio-Temporal Contrastive Learning Network, feature selection using Equilibrium Optimizer module, and predict the yield using Hierarchical Yield Prediction Module.
MTMS-YieldNet integrates spatial-spectral features with spatio-temporal dependencies through a unified architecture, enabling the model to focus on combined spatial-spectral-temporal representations for enhanced crop yield estimation. 
The framework incorporates a Spectral-Spatial Attention Module for rich spatial-spectral features extraction by utilizing multi-spectral imagery,  enhancing its ability to differentiate crop health and growth variations. To capture long-term temporal dependencies, we introduce the Spatio-Temporal Dependency Module, which applies attention mechanisms to temporal data, enabling the model to capture critical spatio-temporal relationships from time-series imagery.
This is crucial for improving the model's performance across varying climatic and seasonal conditions. 
Also, a Spatio-Temporal Contrastive Learning Network, a novel pre-training approach, proposes a method that requires no external supervision, enabling more efficient learning. 
To preserve essential features, we implement feature selection using the Equilibrium Optimizer module, Improving the model's resilience. The hierarchical structure of MTMS-YieldNet predicts the yield using the Hierarchical Yield Prediction Module, effectively synthesizing spatial, spectral, and temporal data, ensuring accurate crop yield estimates. This comprehensive approach empowers MTMS-YieldNet to provide consistent and precise yield predictions, even in complex agricultural environments. This research provides the following key insights:
\begin{itemize}
    \item We propose MTMS-YieldNet as a novel deep learning framework that integrates spatial-spectral features with spatio-temporal dependencies to improve yield prediction. This framework supports better decision-making and resource management by capturing complex patterns in diverse climatic and seasonal conditions.
    \item We present a Self-Supervised Spectral-Spatial-Temporal Feature Extraction framework designed to enhance crop yield prediction by leveraging multi-temporal and multi-spectral satellite imagery. 
     \item We introduce a Spatio-Temporal Contrastive Learning (STCL) Network, a novel pre-training technique that improves the model's learning efficiency by leveraging contrastive learning principles. 
     \item We incorporate the Equilibrium Optimizer (EO) module for feature selection to refine features and enhance the performance of our proposed framework.  
      \item We developed a Hierarchical Yield Prediction Module that combines spatial, spectral, and temporal data to enhance yield estimation accuracy by capturing complex patterns.
\end{itemize}

This research is structured in the following sections: Section~\ref{sec: related} reviews the existing literature on crop yield prediction models and outlines the challenges addressed by our proposed framework. Section~\ref{sec: sa_Prob} explains the problem statement with the scope of our proposed approach. Section~\ref{sec: method} details the design and implementation of our model, including its components and the integration of various modules. Section~\ref{sec: exp} provides the details of the experimental setup,
including datasets, metrics, and results, comparing our methodology with SOTA models.  
The discussion for seasons and years is described in Section~\ref{sec: Dis}. 
Finally, Section~\ref{sec: conclude} summarizes the paper and highlights the contributions of our framework to crop yield prediction and agricultural management.

\section{Related work}
\label{sec: related}
In this section, we provide a detailed overview of recent studies and highlight the key challenges in agricultural yield prediction.
To facilitate better understanding, we organize yield prediction techniques into three categories: Traditional yield prediction methods, Deep learning-based yield prediction methods, and Attention-enhanced and spatio-temporal deep learning methods.

\subsection{Traditional yield prediction methods}
In the last decades, crop yield prediction methods have utilized vegetation index (VI) data in various ways by including the normalized difference vegetation index and the enhanced vegetation index \citep{Shuai_2022}. 
Machine learning methods used in yield prediction perform relative analyses between vegetation indices and crop yields.  Various methods utilized spatio-temporal information efficiently including multi-layer regression \citep{Piekutowska_2021}, and support vector regression \citep{Zhang_2020}, but relied on a limited number of spectral bands. 
Additionally, these approaches do not provide information for the spatial variability within the images and struggle to extract meaningful spatial-spectral features. Traditional approaches struggle to deal with multi-spectral images and face difficulties in managing variations in spatial information for crop yield.

\subsection{Deep learning-based yield prediction methods}
With the emergence of deep learning, the limitations of conventional machine learning approaches are overcome by automatically discovering spectral features and utilizing all available spatial-spectral features in multi-spectral images (MSIs). Convolutional neural networks (CNNs) are used to identify spatial structure in images by analyzing neighboring patches. These neural networks are used as core technology to extract spatial features and spectral features related to agricultural patterns, and they have remarkable performance in agricultural applications such as urban planning and precision farming \citep{Yuan_2020}. 
Also, other methods incorporate transfer learning 
and propose deep learning architectures that combine spatial and spectral information \citep{Yang2019}.
These methods improve yield estimation by capturing spatial-spectral patterns with 2D-CNNs but struggle to process three-dimensional data cubes that include spatio-spectral details in MSIs.
Some studies have introduced 3D CNNs to tackle this limitation by simultaneously considering both spectral and spatial dimensions \citep{Qiao_2021}. 
Unlike 2D CNNs, 3D CNNs use three-dimensional convolution kernels to capture spectral correlations across adjacent bands through enhanced feature extraction. However, they fail to capture the relative analysis of spectral-spatial features and fine-grained spatial details within individual bands. Deep Learning methods have also been explored for other agriculture-based tasks, such as query-response systems~\citep{Rehman_2023} to help farmers in decision-making to improve crop production. 

\subsection{Attention-enhanced and spatio-temporal deep learning methods}\
Attention-based mechanisms emphasize key features or regions in an image to improve prediction performance. 
This mechanisms were first introduced in machine translation application \citep{NIPS2017_3f5ee243} of natural language processing by highlighting relevant information and filtering out the less relevant data.
 Recently, researchers have applied attention-based techniques to hyperspectral image classification along with spatial-spectral analysis within convolution-based models \citep{Pu_2021}. Multi-attention models have further advanced this approach by selectively amplifying important features
 \citep{Raghaw_2025, Qu_2024}. 
 \cite{Ahmad_2024} developed SENet with channel attention for feature map adjustment, while other attention-based models include the Transformer, Spatial Transformer Network, and ACNN that are used in tasks like object recognition.
 In the agricultural sector, attention mechanisms have proven highly effective for crop disease detection along with yield prediction and plant phenotyping.
The latest advancement in crop yield prediction by developing various models through deep learning techniques for extracting spatiotemporal features, which have significantly improved prediction performance. Compared to traditional statistical models, deep learning methods by including Convolutional Neural Networks (CNNs) along with Recurrent Neural Networks (RNNs) have achieved better performance.
In recent years, Convolutional LSTM (ConvLSTM) networks have been used for spatio-temporal tasks because they capture spatial features using CNNs and temporal dependencies at the same time using LSTMs. ConvLSTMs have proven effective for applications in vineyard yield prediction with spatio-temporal feature extraction \citep{Kamangir_2024}.
However, these models struggle with retaining long-term spatio-temporal dependencies over time \citep{zhou_2024}. To improve model performance, researchers have applied various augmentation techniques. 
These techniques assist in generating synthetic data for expanding training datasets, which is beneficial when real-world data remains limited or contains noise.  
Researchers have also combined GANs with attention mechanisms to refine agricultural image augmentation for improving prediction performance in variable environmental conditions \citep{Lu_2022}.
Spatio-temporal fusion (STF) methods combine remote sensing images with varying spatio-temporal resolutions to improve monitoring frequency and accuracy. Various STF techniques have achieved notable success. However, by depending on linear assumptions, traditional methods struggle to manage complex and dynamic scenes.
In contrast, through handling complex scenarios, diffusion-based spatio-temporal fusion methods have outperformed GAN-based approaches by improving precision in dynamic environments \citep{Huang_2024, Song2021, Ho2020}. Researchers have also investigated self-supervised learning for crop yield prediction. This method helps models extract relevant features from large unlabeled data before fine-tuning them for specific tasks. It leverages multi-modal spatio-temporal data to enhance agricultural precision by capturing complex dependencies \citep{Lin_2023}.

In summary, current yield prediction methods face challenges in integrating spatial-spectral and temporal data. 
Variability in agricultural landscapes creates difficulties related to spatial information while spectral limitations make identifying crop characteristics difficult. Temporal changes in environmental conditions and crop growth also complicate the prediction process.

\section{Problem formulation}
\label{sec: sa_Prob}
Consider \( D \) as a dataset with \( N \) samples, where each sample \( {x}_i \) represents multi-temporal and multispectral satellite imagery. Each sample is defined by \( {x}_i \in \mathbb{R}^{T \times H \times W \times C} \), with \( T \) indicating the number of temporal instances, \( H \) and \( W \) representing the spatial dimensions, and \( C \) denoting the count of spectral bands. The objective is to predict crop yield, represented by \( {y}_i \in \mathbb{R}^{d} \), where \( d \) refers to the number of yield parameters for the \( i \)-th sample.
\begin{equation}
{y}_{i} =  f_{SST}(\mathit{{x}}_i)
\label{eq: Prob}
\end{equation}
We propose a method \( f_{\text{SST}} \) designed for efficient extraction of spectral, spatial, and temporal features from the satellite imagery \( {X} \), where \( {x}_i \in {X} \). The method outputs a predicted yield \( Y_{\text{pred}} \), such that \( {y}_i \in Y_{\text{pred}} \), is described by Eq.~(\ref{eq: Prob}). This formalization allows the model to effectively capture complex spatio-temporal dependencies with spectral-spatial correlation thus enhancing the performance and robustness of yield predictions. By utilizing these multi-spectral features, the proposed method aims to improve crop yield estimation across various agricultural conditions.
This research seeks to achieve the following objectives:
\begin{itemize}
    \item To assess the effectiveness of using spatial, spectral, and temporal information from multi-temporal and multi-spectral satellite imagery for accurate crop yield prediction.
    \item To analyze self-supervised contrastive learning for enhancing satellite imagery feature representation.
    \item To explore how attention mechanisms can enhance feature learning from the imagery data.
    \item To investigate the impact of temporal dependencies in the satellite imagery data on crop yield predictions.
\end{itemize}

\begin{figure}[h!]
    \centering
    \includegraphics[width=\linewidth]{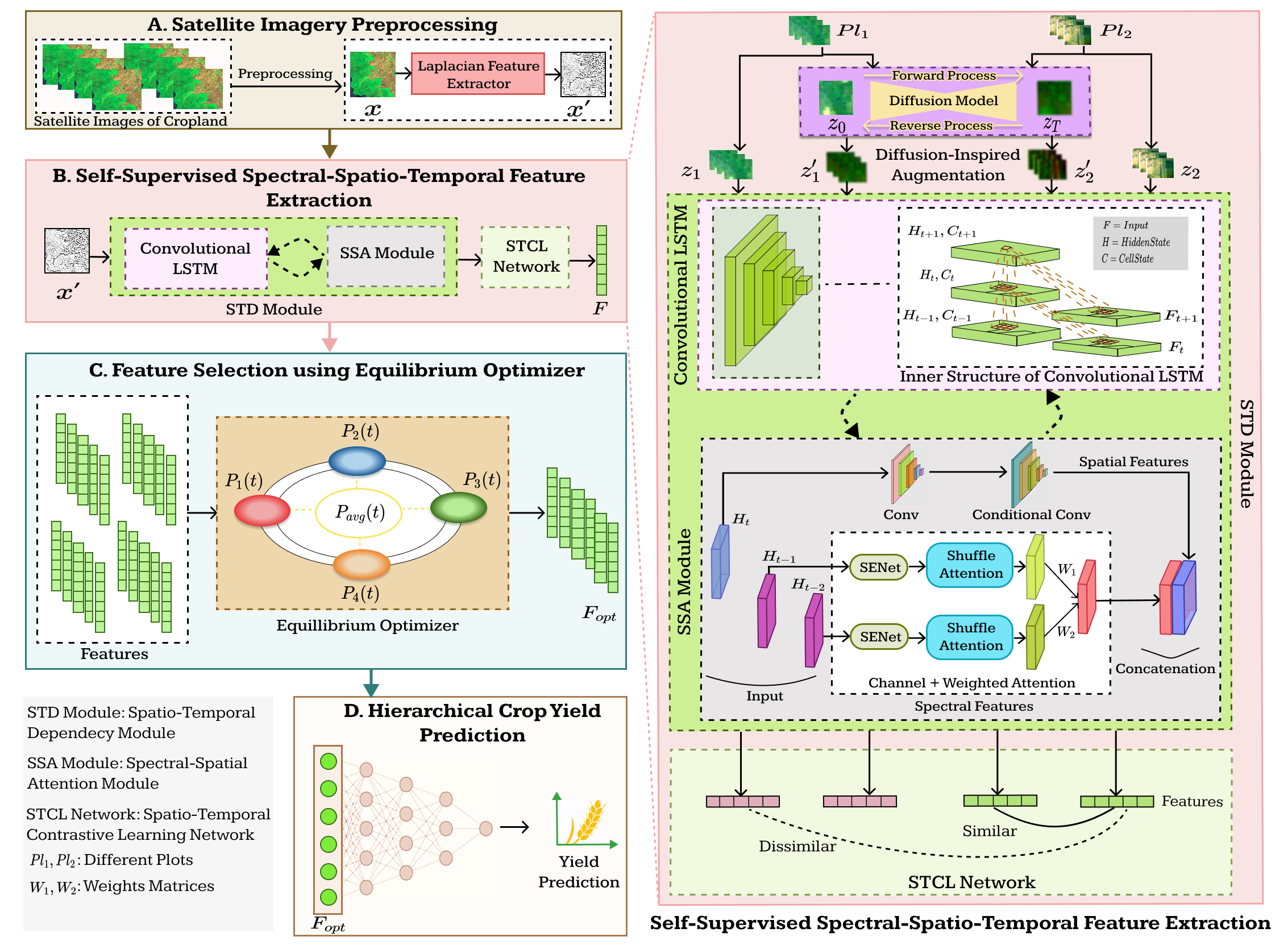}
    \captionsetup{justification=justified, singlelinecheck=false}
    \caption{The architecture of our proposed MTMS-YieldNet consists of the following modules: (A) Satellite imagery preprocessing for feature enhancement; (B) Self-supervised spectral-spatial-temporal feature extraction module extract features from spatial, spectral, and
    temporal data by utilizing contrastive learning technique; (C) Feature selection using equilibrium optimizer based on physics-based particle theory; and (D) Hierarchical yield prediction network utilizing the optimized features and giving prediction map.}
    \label{fig: framework}
\end{figure}

\section{Methodology}
\label{sec: method}
This section provides a thorough analysis along with the overall architectural description of our proposed MTMS-YieldNet framework for yield prediction, as illustrated in Fig. \ref{fig: framework}. The proposed framework has been divided into four different modules:
(A) satellite imagery preprocessing, which improves the quality of raw satellite images by enhancement technique; (B) self-supervised spectral-spatial-temporal feature extraction extracts features from spatial, spectral, and temporal data using self-supervised learning techniques; (C) feature selection using equilibrium optimizer selects the most relevant features by optimizing feature importance to improve prediction accuracy; (D) hierarchical deep learning based yield prediction utilizes a hierarchical deep learning model to predict crop yield by combining the selected features at multiple levels.

\subsection{\textit{Satellite imagery preprocessing}}
To improve the quality of remote sensing data for agricultural forecasting, satellite image preprocessing is important. This process corrects distortions, reduces noise, and standardizes data for consistency. 
To enhance contrast and emphasize edges and spatial structures, which are important for diagnosing crop health, a  Laplacian filter is applied to an image \( x \).
This filter emphasizes edges and boundaries, effectively highlighting spatial structures crucial for assessing crop health and predicting yield by capturing local intensity changes in the spatial dimensions \( H \) and \( W \). After preprocessing, we got \( x' \), that is a preprocessed image which is considered as input for the next feature extraction module known as Self-Supervised Spectral-Spatial-Temporal Feature Extraction. In this module, the preprocessed image is passed to the spatio-temporal dependency module (STD module). This module helps to extract the features that are refined by contrastive learning based on similarities and dissimilarities between the features.

\subsection{\textit{Self-supervised spectral-spatial-temporal feature extraction}}
The scarcity of labeled data, seasonal variations, environmental noise, and the complexity of dynamic growth patterns present significant difficulties in crop yield prediction. Self-supervised contrastive learning provides a solution by allowing the model to learn robust features without needing extensive labeled data, helping distinguish between similar and dissimilar patterns. This module used the potential of self-supervised learning to capture intrinsic features from satellite image datasets. For the analysis and understanding of crop health with growth stages, and environmental impacts, this module helps extract the key features from remote sensing datasets including spatial patterns, spectral characteristics and temporal variations. The pre-training process is categorized into three submodules: the Diffusion-Inspired Augmentation, which enhances data variability and robustness; the Spatio-Temporal Dependency Module, which captures relationships over long timeseries data and spatial-spectral features from multi-spectral imagery; and the Spatio-Temporal Contrastive Learning Network, which enables model training without relying on extensive supervision.

\subsubsection{\textit{ Diffusion inspired augmentation}}
Selecting an appropriate augmentation technique is essential in contrastive learning as it significantly enhances the diversity and quality of the training data. 
Traditional Generative Adversarial Network (GAN) based approach have been commonly used for data augmentation. However, they present several challenges, including difficulties in generating high-quality variations, the risk of distribution collapse, and issues with maintaining the consistency of spatial and temporal features. To address these challenges, diffusion-based augmentation emerges as a more effective technique, generating high-quality variations of the original data through a process of forward and reverse diffusion. 
The diffusion-based image augmentation module generates synthetic images by utilizing the principles of the Trajectory Consistency Distillation (TCD) Scheduler. 
In Fig. \ref{fig: framework}, we have taken images $z_1$ and $z_2$ from different plots, respectively, $Pl_1$ and $Pl_2$. Following this, we applied an augmentation task using the diffusion model, resulting in the augmented images $z'_1$ and $z'_2$.
\begin{figure}[h!]
    \centering
    \captionsetup{justification=centering}
    \includegraphics[width=10cm, height=5cm]{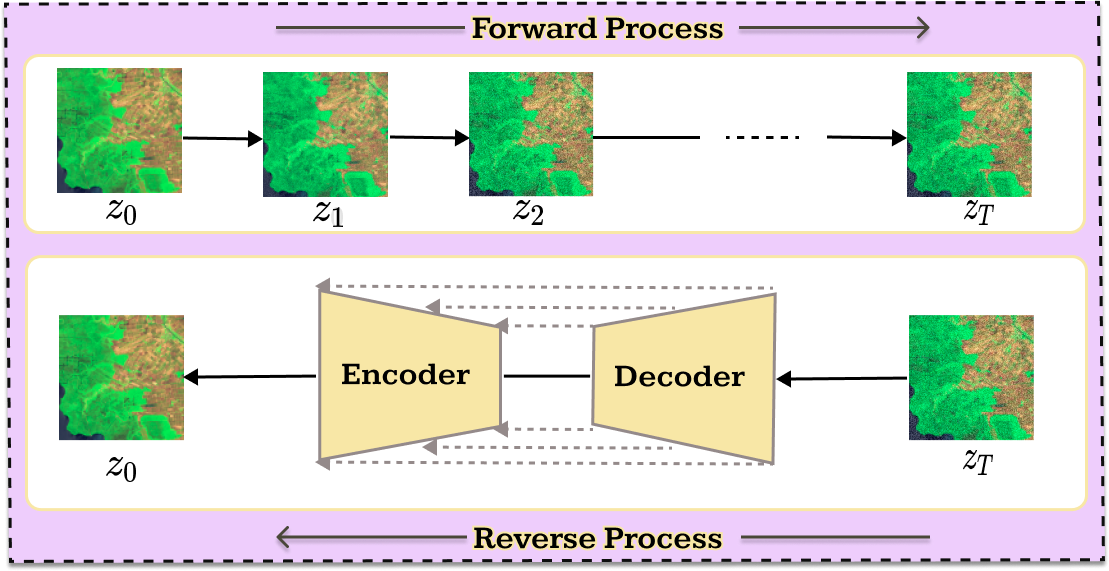}
    \caption{Diffusion-inspired augmentation using forward and reverse process. }
    \label{fig: Diffusion}
\end{figure}
This process involves the following steps: The forward diffusion progressively adds noise to the images over time, gradually corrupting them and creating diverse training samples. This is followed by the reverse diffusion 
step, where a generative network aims to reconstruct clean
images from the noisy data by predicting the original image from its noisy counterpart. To ensure that the predicted trajectory of the generative process remains consistent throughout the reverse diffusion process, the Trajectory Consistency Function is employed, maintaining high-quality image
generation. Finally, Strategic Stochastic Sampling (SSS) optimizes the noise addition and removal process to minimize error accumulation across multiple steps, thereby enhancing the overall quality and stability of the augmented images. The major components of diffusion are shown in Fig. \ref{fig: Diffusion}. The forward diffusion process gradually adds noise to an image over time, where the noisy image at time \( t \), denoted as \( z_t \), is defined by the Eq.~(\ref{eq: FD}). In this equation, \( z_0 \) is the original image, \( \beta_t \) controls the noise schedule, and \( \mathcal{N} \) represents the Gaussian distribution. This process generates a series of progressively noisy images, enriching the training dataset with diverse variations that help the model learn robust features.
\begin{equation}
\mathit{q}({z}_t \mid {z}_0) = \mathcal{N}({z}_t; \beta_t {z}_0, (1 - \beta_t){I}) \tag{2}
\label{eq: FD}
\end{equation}

\begin{equation}
\mathit{p}_\theta({z}_{t-1} \mid {z}_t) = \mathcal{N}({z}_{t-1}; \mu_\theta({z}_t, t), \sigma_t^2 {I})  \tag{3}
\label{eq: p_theta}
\end{equation}

The reverse diffusion process reconstructs the clean image \( z_0 \) from its noisy counterpart \( z_t \) by iteratively denoising the images. This process is modeled as a generative process, where the conditional probability of transitioning from state \( z_t \) to \( z_{t-1} \) is given in Eq.~(\ref{eq: p_theta}), where \( \mu_\theta(z_t, t) \) represents the predicted mean, and \( \sigma_t \) denotes the variance at time \( t \). The model predicts the previous state \( z_{t-1} \) by estimating \( \mu_\theta(z_t, t) \) and adding controlled noise, iteratively reconstructing \( z_0 \). 
To ensure consistency with the noisy trajectory, the trajectory consistency function measures the deviation between the predicted images \( G_\theta(z_0, t) \) and the noisy images \( z_t \), as shown in Eq.~(\ref{eq: consistency}), where \( G_\theta(z_0, t) \) is the generator function at time \( t \), and \( \| \cdot \|_2 \) represents the L2-norm. Minimizing \( T_{\text{consistency}} \) assures that the generated images align with the noisy trajectory, preserving high fidelity and accuracy in the reconstructed images.

\begin{equation}
\mathit{T}_{\text{consistency}}(\mathbf{z}_0, t) = \| \mathbf{z}_t - G_\theta(\mathbf{z}_0, t) \|_2^2   \tag{4}
\label{eq: consistency}
\end{equation}

 \begin{equation}
\mathit{L}_{\text{SSS}} = \sum_t \mathcal{L}_{\text{reg}}(\mathbf{z}_t, \mathbf{z}_0) + \lambda T_{\text{consistency}}(\mathbf{z}_0, t)  \tag{5}
\label{eq: sss}
\end{equation}

Strategic Stochastic Sampling (SSS) optimizes the noise addition and removal process to reduce error accumulation and enhance the stability of augmented images. Eq.~(\ref{eq: sss}) described the objective of SSS optimization in which \( L_{\text{reg}}(z_t, z_0) \) is the regularization loss between the original image \( z_0 \) and noisy image \( z_t \), and \( \lambda \) controls the weight of the trajectory consistency. This method is helpful to minimizing the \( L_{\text{reg}} \) to preserve proximity to the original images with maintaining \( T_{\text{consistency}} \)  consistent image generation. The model achieves a balance between fidelity and diversity by adjusting \( \lambda \), leading to more stable and accurate image augmentation.

\subsubsection{\textit{Spatio-temporal dependency module}}
\begin{figure}[h!]
    \centering
    \captionsetup{justification=centering}
    \includegraphics[width=8.5cm, height=3.5cm]{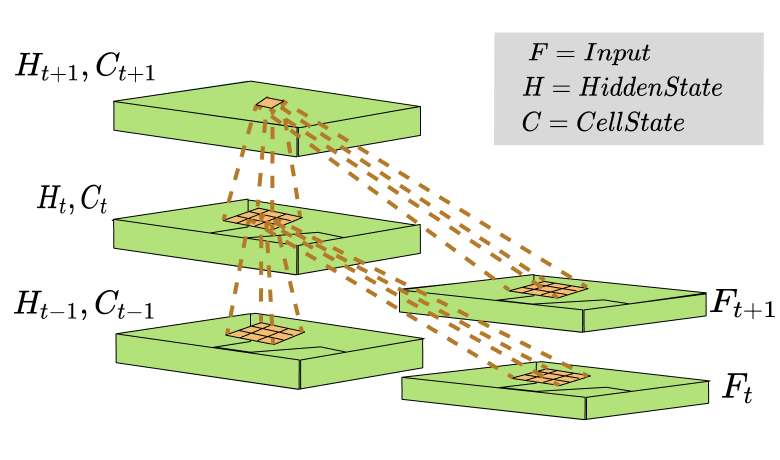}
    \caption{Inner structure of ConvLSTM. }
    \label{fig: ConvLSTM}
\end{figure}

The variability in agricultural landscapes along with spectral and temporal challenges presents significant obstacles to accurately predicting crop yields. Our model introduces a new Spatio-Temporal Dependency (STD) module to address these issues that detects spatio-temporal dependencies using ConvLSTM and spatial-spectral patterns with a Spectral-Spatial Attention (SSA) module. ConvLSTM is composed of interconnected convolutional and recurrent units that help the model to preserve long-term temporal dependencies while preserving spatial hierarchies, as shown in Fig.~\ref{fig: ConvLSTM}.  

The SSA module integrates with the ConvLSTM layer to form a unified framework for capturing correlations of spatio-temporal features. The SSA module improves the feature extraction abilities of ConvLSTM with utilizing the potential of attention mechanisms and cascaded convolutions and addressing the challenges associated with long-term dependency loss. 

This integration provides a strong representation of time-series data is essential for precise crop yield predictions. Traditional models inspire the ConvLSTM layer and capture long-term temporal and spatial dependencies using convolutions instead of matrix multiplications for input-to-state and state-to-state transformations. The ConvLSTM contains unique characteristics and handles spatial features with temporal dependencies. It contains three gates: input (\( i_t \))  Eq.~(\ref{eq: 6a}), forget (\( f_t \))  Eq.~(\ref{eq: 6b}), and output (\( o_t \)) Eq.~(\ref{eq: 6d}), which manage data flow and ensure the efficient transmission of relevant information. These gates help in the efficient representation of spatio-temporal features by using multidimensional convolutional operations instead of one-dimensional ones.  

\begin{equation}
\mathit{i}_t = \sigma(W_{Fi} \cdot {F}_t + W_{Hi} \cdot {H}_{t-1} + W_{Ci} \odot {C}_{t-1} + b_i)  \tag{6a}
\label{eq: 6a}
\end{equation}
In Eq.~(\ref{eq: 6a}), \( {F}_t \) is the input at time \( t \), \( {C}_{t-1} \) is the cell state, and \( {H}_{t-1} \) represents the hidden state. The variables \( i_t, f_t, \) and \( o_t \) represent the input, forget, and output gates, respectively. The symbol \( \cdot\) denotes the convolution operation, \( \odot \) refers to the Hadamard product, the kernel size of \( k \times k \) for the convolution filters is denoted by \( W \), and the bias for each gate is represented by \( b \).

\begin{equation}
\mathit{f}_t = \sigma(W_{Ff} \cdot {F}_t + W_{Hf} \cdot {H}_{t-1} + W_{Cf} \odot {C}_{t-1} + b_f)  \tag{6b}
\label{eq: 6b}
\end{equation}
\begin{equation}
{C}_t = \mathit{f}_t \odot {C}_{t-1} + \mathit{i}_t \odot \tanh(W_{Fc} \cdot {F}_t + W_{Hc} \cdot {H}_{t-1} + b_c)  \tag{6c}
\label{eq: 6c}
\end{equation}
For the previous cell state \( {C}_{t-1} \), the weight matrics are \( {W}_{Ci} \) to input gate, \( {W}_{Cf} \) to forget gate, and \( {W}_{Co} \) to output gate.
In Eq.~(\ref{eq: 6c}), the cell state \( {C}_t \) is updated by combining the previous cell state with the forget gate and input gate regulating the new candidate values.

\begin{equation}
\mathit{o}_t = \sigma(W_{Fo} \cdot {F}_t + W_{Ho} \cdot {H}_{t-1} + W_{Co} \odot {C}_t + b_o)  \tag{6d}
\label{eq: 6d}
\end{equation}

\begin{figure}[h!]
    \centering
    \captionsetup{justification=centering}
    \includegraphics[width=0.6\textwidth]{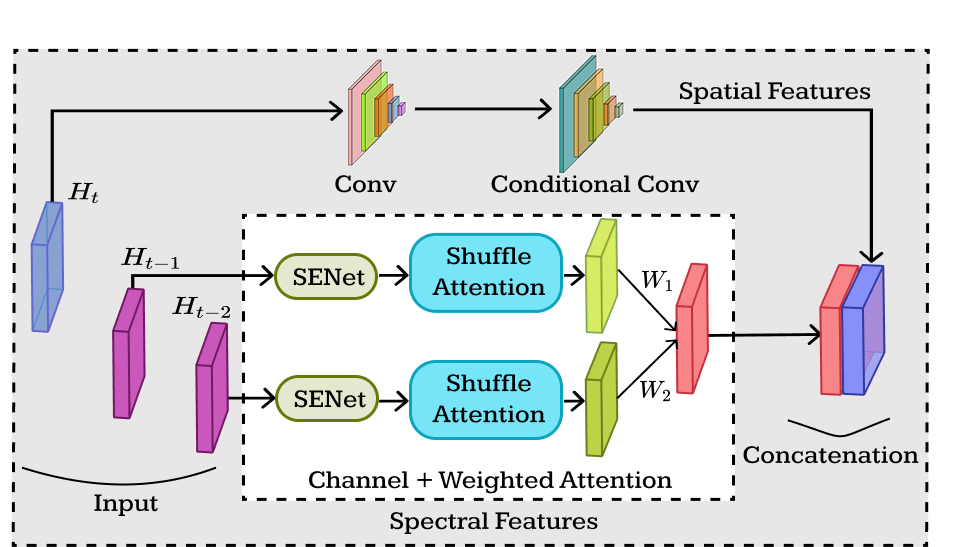}
    \caption{Architecture of spectral-spatial attention module.}
    \label{fig: SSA}
\end{figure}
 
In Eq.~(\ref{eq: 6e}), the hidden state \( {H}_t \) is derived by applying the output gate to the \( tanh \) of the cell state. The weight matrices \( {W}_{Fi} \), \( {W}_{Ff} \), \( {W}_{Fo} \), and \( {W}_{Fc} \) correspond to the input \( {F}_t \) with respect to the input gate, forget gate, output gate, and candidate cell state. Similarly, \( {W}_{Hi} \), \( {W}_{Hf} \), \( {W}_{Ho} \), and \( {W}_{Hc} \) are weight matrices for the previous hidden state \( {H}_{t-1} \) with respect to the input gate, forget gate, output gate, and candidate cell state.
 
\begin{equation}
{H}_t = \mathit{o}_t \odot \tanh({C}_t)   \tag{6e}
\label{eq: 6e}
\end{equation}

\begin{figure}[h!]
    \centering
    \captionsetup{justification=centering}
    \includegraphics[width=\textwidth]{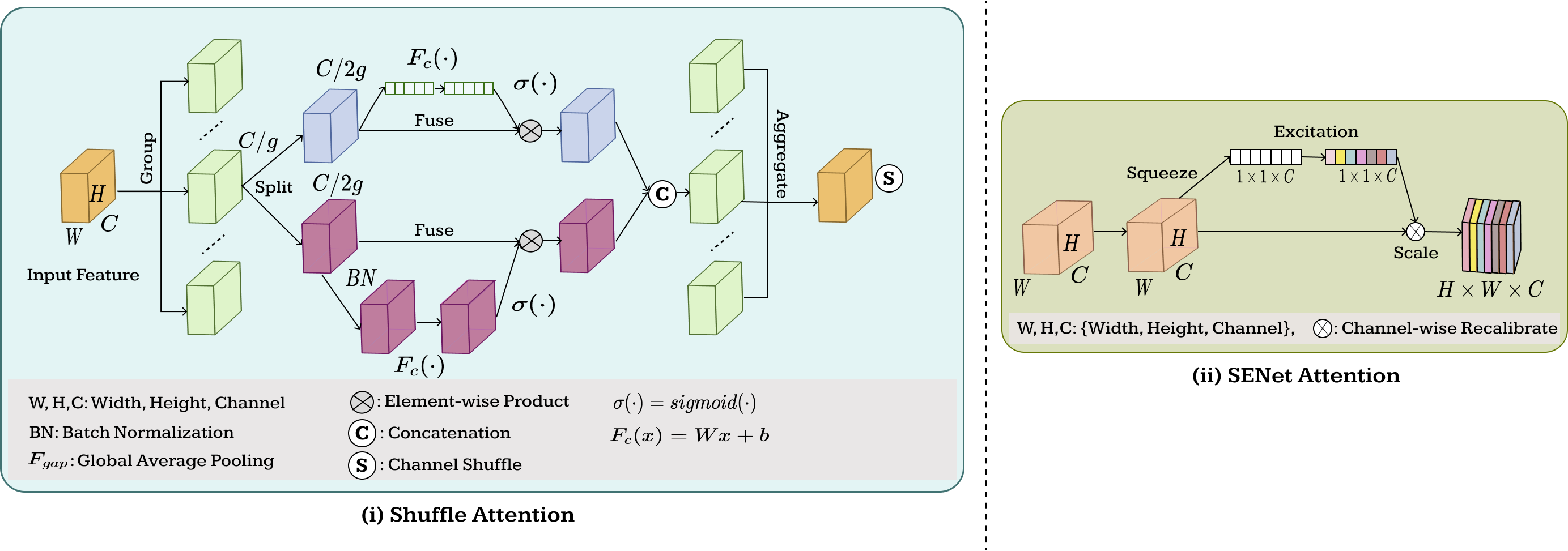}
    \caption{Overview of applied attention mechanism: (i) Shuffle attention; (ii) SENet attention.}
    \label{fig: shuffle}
\end{figure}

To extract spectral-spatial feature correlation, we propose a Spectral-Spatial Attention (SSA) module, as shown in Fig.~\ref{fig: SSA}. A specialized convolutional network that integrates stacked attention mechanisms 
with cascaded conditional and standard convolutional layers. This architecture captures spatial features from each image in the sequence, enhancing spatial representation. 
The SSA module employs a cascaded architecture for the extraction of spectral features \( {H}_{t-1:t-a} \), integrating multiple attention mechanisms to enhance feature representation. Initially, the Squeeze-and-Excitation Network (SENet) attention block \citep{Ahmad_2024} adaptively models channel relationships by recalibrating feature responses, emphasizing the most informative channels while suppressing less relevant ones, as illustrated in Fig.~\ref{fig: shuffle}(ii). This process, defined in Eq.~(\ref{eq: SE}), involves a combination of global average pooling, learnable weight matrices  \( W_1 \) and \( W_2 \), ReLU activation \( \delta \), and a sigmoid activation \( \sigma \), followed by element-wise multiplication \(\odot\).

To further enhance cross-channel information flow, channel shuffle attention is applied, as formulated in  Eq.~(\ref{eq: Shuffle}). This mechanism illustrated in Fig.~\ref{fig: shuffle}(i) involves channel grouping and shuffling, enhancing robustness to variations in channel-wise information and facilitating effective information exchange. Lastly, weighted temporal attention is incorporated to balance the influence of sequential features over time. According to Eq.~(\ref{eq: Temp}), this process assigns learnable weights \( w_\tau \in \mathbb{}{R} \) to balance temporal feature contributions, ensuring that the model effectively captures long-range dependencies while preserving spectral integrity.

\begin{equation}
\mathit{SE}({H}_{t-\tau}) = \sigma(W_2 \delta(W_1 \mathit{Pool}({H}_{t-\tau}))) \odot {H}_{t-\tau}  \tag{7}
\label{eq: SE}
\end{equation}

\begin{equation}
\mathit{ShuffleAtt} ({H}_{t-\tau}) = \mathit{Shuffle}(\mathit{ChannelGroup} (\mathit{SE}({H}_{t-\tau}))) \tag{8}
\label{eq: Shuffle}
\end{equation}

\begin{equation}
\mathit{TempAtt} ({H}_{t-\tau}) = \sum_{\tau=t-a}^{t-1} w_\tau \, \mathit{ShuffleAtt} (\mathit{SE}({H}_{t-\tau})) \tag{9}
\label{eq: Temp}
\end{equation}

The spatial feature extraction begins with conditional and standard convolutional layers, which capture spatial details from each image in the sequence. 
These layers are followed by stacked attention modules, allowing the model to concentrate on important spatial regions. For spatial features \( {H}_t \), a series of standard convolutions followed by conditional convolutions are applied, as shown in Eq.~(\ref{eq: Conv}). In this case, \(\text{CondConv}\) dynamically generates filter weights based on the input features, as defined in Eq.~(\ref{eq: CondConv}).

\begin{equation}
\mathit{ConvStack} ({H}_t) = \mathit{CondConv}(\mathit{Conv}({H}_t)) \tag{10}
\label{eq: Conv}
\end{equation}

\begin{equation}
\mathit{CondConv}({x}) = \sum_{k} \pi_k({x}) \odot W_k \odot {x}  \tag{11}
\label{eq: CondConv}
\end{equation}

The dynamic filtering mechanism uses \( \pi_k(\mathbf{x}) \) to represent data-dependent routing weights and \( W_k \) is a representation of filter banks. This enable the model to adaptively apply filters to different input instances to improve its ability to capture complex spatial patterns. 
In the next step feature concatenation is performed by combining the outputs from spectral and spatial components. 
This process combines spatially refined spectral features extracted through a convolutional stack (\( \text{ConvStack}(H_t) \)) and temporally focused spectral features derived from a series of previous time steps using a Temporal Attention Mechanism (\( \text{TempAtt}(H_{t-a:t-1}) \)). The spectral component captures localized spectral patterns within the input tensor \( H_t \) and the spatial component highlights spatial dependencies across the sequence \( H_{t-a} \) to \( H_{t-1} \). These outputs are concatenated along the feature dimension to create a unified representation that includes both spatial and spectral information, as shown in Eq.~(\ref{eq: Concat}).

\begin{equation}
F_{concat} = \left[ \mathit{ConvStack}(H_t), \mathit{TempAtt}(H_{t-a:t-1}) \right] \tag{12}
\label{eq: Concat}
\end{equation}

This unified representation enhances the model’s ability to effectively process spatial-spectral dependencies with making it highly suitable for tasks such as crop yield prediction.
 
\subsubsection{\textit{Spatio-temporal contrastive learning network}}
Self-supervised learning (SSL) has introduced as a powerful mechanism for pre-training models without needing external human intervention. This work is also leverages with SSL to train our Spatio-Temporal Dependency (STD) module for time-series satellite imagery and using a strength of contrastive loss to learn complex spatio-temporal representations. In this context, the key idea behind self-supervised approach is to generate pair of images as positive and negative pairs of image sequences. Positive pairs are the augmented versions of the same image, while negative pairs are formed by comparing images from different plots as well as within the same plot but captured at different timestamps. 
We use diffusion-based augmentation to achieve this, which generates diverse and realistic transformations of the original sequence. This ensures that the model learns to focus on the meaningful features of the data while being invariant to unimportant changes introduced by the augmentation process. The contrastive loss is defined by Eq.~(\ref{eq: contrastive}):

\begin{equation}
\mathit{L}_{\text{contrastive}} = -\log \left( \frac{\exp\left(\text{sim}({v}_t^{(1)}, {v}_t^{(2)}) / \tau \right)}{\sum_{t' \neq t} \exp\left(\text{sim}({v}_t^{(1)}, {v}_{t'}^{(2)}) / \tau \right) + \exp\left(\text{sim}({v}_t^{(1)}, {v}_t^{(2)}) / \tau \right)} \right)  \tag{13}
\label{eq: contrastive}
\end{equation}
where, \( {v}_t^{(1)} \) and \( {v}_t^{(2)} \) are the final representations (classification tokens) for the positive pair derived from the augmented sequences  $z'_1$ and $z'_2$, 
\( \text{sim}({v}_i, {v}_j) \) represents the cosine similarity between vectors \( {v}_i \) and \( {v}_j \),
\( \tau \) is the temperature parameter that controls the sharpness of the similarity distribution.
By optimizing this contrastive loss, the model learns to capture consistent spatio-temporal features across different augmentations, effectively learning useful representations without external supervision. 

\subsection{\textit{Feature selection using equilibrium optimizer}}
\begin{figure}[h!]
    \centering
    \captionsetup{justification=centering}
    \includegraphics[width=0.8\textwidth]{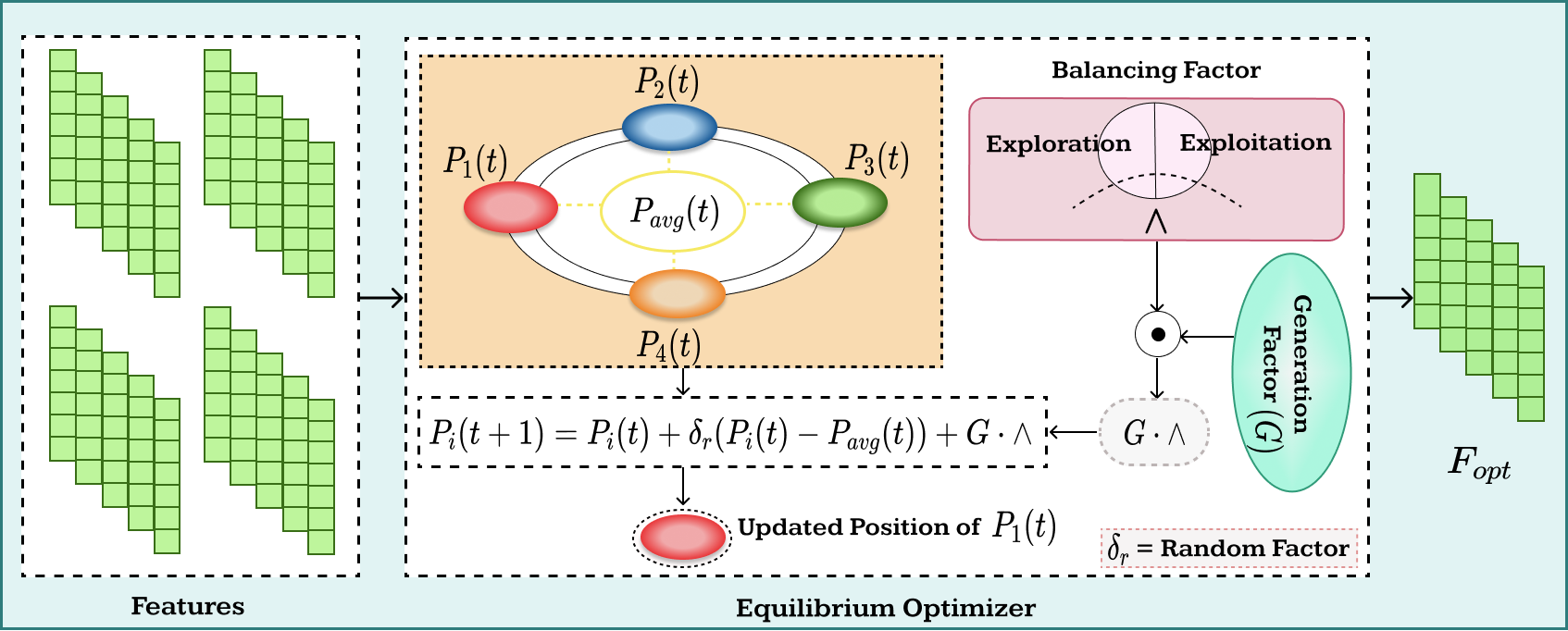}
    \caption{Feature selection through equilibrium optimizer.}
    \label{fig: EO}
\end{figure}

The Equilibrium Optimizer (EO) algorithm is inspired by the principles of physics-based particle theory and is designed to optimize the feature selection process, as described in Fig.~\ref{fig: EO}. This method enhances the feature extraction and refinement by iteratively exploring and exploiting the feature space, ensuring that only the most significant features are selected. In the context of the MTMS-YieldNet model, EO optimizes feature selection for crop yield prediction using multi-spectral and multi-temporal satellite imagery.

The EO algorithm initializes a set of particles. Each particle represents a potential solution in the feature space, where the position of a particle corresponds to the selection of a subset of features. Let \( P_{i,j} \) represent the position of the \( j \)-th dimension of the \( i \)-th particle, where \( P_{i,j} \) is either 0 or 1, indicating whether the feature is selected or not. The position of the particles at initialization is given by Eq.~(\ref{eq: Particle}):

\begin{equation}
P_{i,j} = P_{i,j} + \delta_r \cdot F   \tag{14}
\label{eq: Particle}
\end{equation}

where, \( \delta_r \) is a random factor between [0, 1], and \( F \) is the feature adjustment vector. This process initializes the particles in a random feature space. After this step, there is a need to calculate the fitness of each particle.
The fitness function evaluates how well each feature subset contributes to the crop yield prediction. Each particle's fitness value is computed based on its selected feature subset. Let the fitness of the \( i \)-th particle at iteration \( t \) be denoted as \( f_i(t) \), calculated by evaluating the model's prediction performance with the selected feature subset.
After the fitness values are calculated, the particles are sorted based on their fitness scores, and the top-ranked particles form the equilibrium pool. The equilibrium pool at iteration \( t \) consists of the four best-ranked particles, denoted as:

\begin{equation}
\mathit{E}(t) = \left[ \mathit{P}_1(t), \mathit{P}_2(t), \mathit{P}_3(t), \mathit{P}_4(t) \right] \tag{15}
\label{eq:15}
\end{equation}
where, \( P_1(t), P_2(t), P_3(t), P_4(t) \) represent the four top performing particles at iteration \( t \) in Eq.~(\ref{eq:15}). The average position of these particles is computed as the global attractor:

\begin{equation}
\mathit{P}_{\text{avg}}(t) = \frac{1}{4} \sum_{i=1}^{4} P_i(t)   \tag{16}
\label{eq:16}
\end{equation}

In Eq.~(\ref{eq:16}), the average position guides the particle search process in subsequent iterations. During each iteration, the position of each particle is updated by considering both local search (exploring around the current position) and global search (exploring the entire feature space). This balance between exploration and exploitation is controlled by the generation factor \( G \) and the attractor \( P_{\text{avg}}(t) \). The position update rule for the \( i \)-th particle is given by Eq.~(\ref{eq:17}):

\begin{equation}
\mathit{P}_i(t+1) = \mathit{P}_i(t) + \delta_r \cdot \left( \mathit{P}_i(t) - \mathit{P}_{\text{avg}}(t) \right) + \mathit{G} \cdot \mathit{\Lambda}   \tag{17}
\label{eq:17}
\end{equation}
where, \( G \) is the generation factor, \( \Lambda \) is the balance parameter between exploration and exploitation, and \( \delta_r \) is the random factor. The generation factor \( G \) is calculated as:

\begin{equation}
\mathit{G} = [\mathit{P}_i(t) - \mathit{P}_i(t-1)] \cdot \mathit{\alpha}         \tag{18}
\label{eq:18}
\end{equation}

\( \alpha \) is a control parameter that tunes the magnitude of the search step as outlined in Eq.~(\ref{eq:18}). The balancing between exploration and exploitation is challenging in the EO algorithm. The parameter \( \Lambda \) maintains the balance between exploration and exploitation and influences the particle’s movement within the feature space. The generation factor \( G \) regulates the exploration ability of the particles, helping to direct the optimization process toward unexplored areas of the feature space. 
The final step in the optimization process selects the best feature subset. The particles adjust their positions to prioritize the most promising solutions and ensure that the MTMS-YieldNet model converges on an optimal set of features. EO reduces overfitting by refining the feature subset iteratively and enhancing computational efficiency.

\subsection{Hierarchical yield prediction network}
\begin{figure}[h!]
    \centering
    \captionsetup{justification=centering}
    \includegraphics[width=0.67\textwidth]{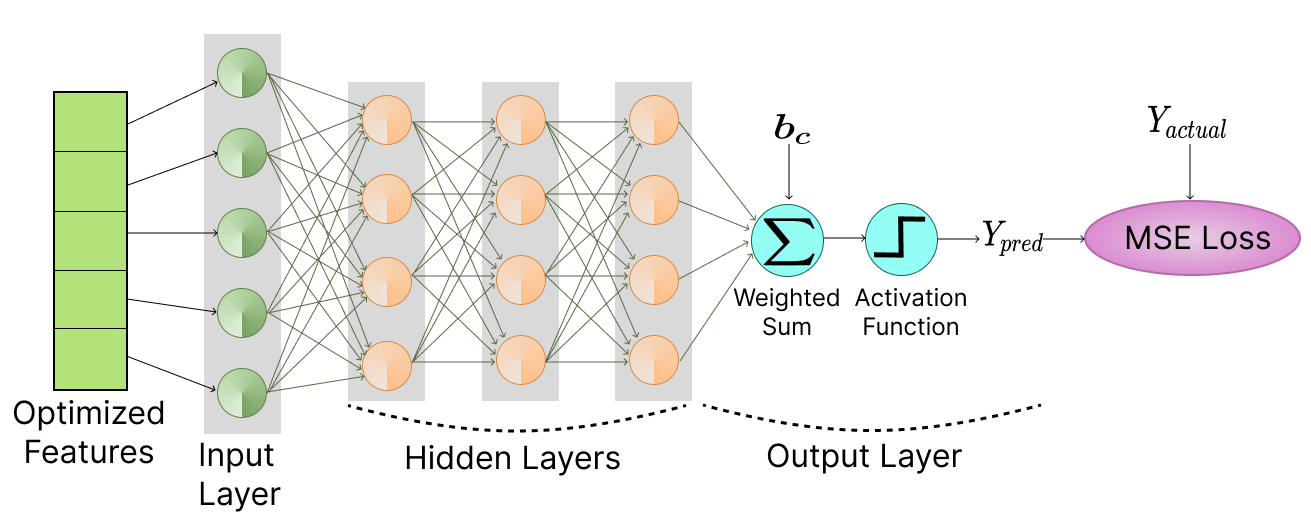}
    \caption{Architecture of Hierarchical yield prediction network.}
    \label{fig: Final_Loss}
\end{figure}

In our proposed framework, the final stage is a hierarchical deep learning-based yield predictor, which utilizes the optimized features \( {F}_{\text{opt}} \) obtained from a particle-theory-based optimization process for enhancing the crop yield estimation, as illustrated in Fig.~\ref{fig: Final_Loss}. This module employs a convolutional neural network-based classification layer with input as optimized features, a kernel size of \( 3 \times 3 \), and padding of 1, facilitating the effective transformation of spatio-temporal features into a structured prediction map. The mathematical representation for the crop yield estimation process is defined by Eq.~(\ref{eq: yield}).

\begin{equation}
Y_{\text{pred}} = \sigma \left( W_c \ast F_{\text{opt}} + b_c \right)         \tag{19}
\label{eq: yield}
\end{equation}

Here, \( W_c \) refers to the convolutional filter, \( * \) denotes the convolution operation, \( b_c \) represents the bias term, and \( \sigma \) is the activation function. 
The Mean Squared Error (MSE) Loss is then computed by comparing the predicted value \( Y_{\text{pred}} \) with the actual ground-truth values \( Y_{i} \), as illustrated in Eq.~(\ref{eq: MSE}):

\begin{equation}
\text{MSE Loss} = \frac{1}{N} \sum_{i=1}^{N} ( Y_{i} - Y_{\text{pred}})^2         \tag{20}
\label{eq: MSE}
\end{equation}
where, N is the number of data points with respect to plots. This structured approach enables the model to efficiently capture spatial-temporal patterns, enhancing crop yield estimation by utilizing effective feature selection. By integrating CNN-based hierarchical learning with optimized feature extraction, the predictor enhances generalization, mitigates overfitting, and ensures robust performance across diverse crop yield conditions.

\section{Experimental settings}
\label{sec: exp}
In this section, we discuss the various datasets employed in our study for the experimental evaluation of the proposed MTMS-YieldNet framework. We also present the experimental setup,  comparative methods and the evaluation metrics of various methods including our proposed approach. Finally, we present the experimental results that highlight the effectiveness of our proposed framework.

\subsection{Dataset summarization}
To evaluate the performance of the proposed approach, we use the SICKLE dataset \citep{Sani_2024}, which includes time-series images from Sentinel-1 (S1), Sentinel-2 (S2), and Landsat-8 (L8). Each of them is considered as a separate dataset for showing the robust analysis. The dataset features multispectral, thermal, and microwave data, along with annotations for important cropping parameters, and covers the Cauvery Delta region in Tamil Nadu, India. The temporal coverage of the dataset from January 2018 to March 2021 enables detailed observation of seasonal crop development patterns. It includes 2,370 seasonal samples from 388 plots, each about 0.38 acres, representing 21 crop types across four districts in the Delta. The dataset contains approximately 209,000 satellite images with a total file size of 6,470 MB and covers a total area of 363.5 hectares. The dataset also includes 351 paddy samples from 145 plots, with details on variety, season, and yield. To ensure a thorough evaluation, the dataset was split into training, validation, and testing sets in an 80-10-10 percent ratio. Table~\ref{tab: dataset} provides a summary of the S1, S2, and L8 datasets including the digitization footprint, which represents the amount of storage required per hectare of land.

\begin{table*}[!ht] \footnotesize
\caption{A detailed summary of the remote sensing datasets utilized in our study, specifying their spatial and temporal resolution with spectral band specification.}
\begin{tabular*}{\linewidth}{@{\extracolsep{\fill}}l ccc}
    \toprule
    {Source} &  {Bands} &  {Digitization Footprint (MB/ha)} \\
    \midrule
    S1  &  VV, and VH  &  2.07\\
    S2  & B1, B2, B3, B4, B5, B6, B7, B8, B8A, B9, B11, and B12 &  6.79\\
    L8  & SR\_B1, SR B2, SR B3, SR B4, SR B5, SR B6, SR B7, and ST B10 & 2.27\\  
    \hline
\end{tabular*}
\label{tab: dataset}
\end{table*}

\textbf{Sentinel-1 (S1)}: The Sentinel-1 dataset comprises microwave images captured by the Sentinel-1 satellite using C-band synthetic-aperture radar (SAR). This dataset includes VV and VH bands with RGB bands assigned to 0, 1, and 0. Its ability to capture images in all weather conditions including both day and night to ensure reliable crop monitoring despite cloud cover.

\textbf{Sentinel-2 (S2)}: The Sentinel-2 dataset includes multi-spectral images with bands B1, B2, B3, B4, B5, B6, B7, B8, B8A, B9, B11, and B12 with RGB bands mapped to 3, 2, and 1, essential for capturing vegetation features and supporting crop monitoring and yield prediction.

\textbf{Landsat-8 (L8)}: The Landsat-8 dataset provides multi-spectral and thermal images containing spectral information SR B1 to SR B7 and ST B10, with RGB bands mapped to 3, 2, and 1. It offers key information on land surface temperature and vegetation health to enhance yield prediction performance.

\subsection{Experimental setup}
This section provides an overview of SOTA methods commonly used as benchmarks to assess the performance of the proposed technique. We also discuss the evaluation metrics applied to analyze different approaches.

\subsubsection{Comparison methods for performance evaluation}
Our presented approach is compared to a wide range of conventional approaches to ensure a thorough assessment. A comprehensive comparison is conducted with the latest techniques for crop yield prediction, including GNN-RNN \citep{Fan_2022},  CNN-GRU \citep{Wang_2023}, UNet \citep{Chakraborty_2023}, 1D-CNN \citep{Khan_2024}, UNet-ConvLSTM \citep{Kamangir_2024}, CTANet \citep{Qu_2024}, and SFC-DenseNet-AM \citep{Seong_2024}, ensuring a rigorous and well-rounded analysis.

\vspace{2mm}
\textbf{GNN-RNN} \citep{Fan_2022} utilized spatio-temporal feature fusion by capturing geographical dependencies among counties using GNN and for sequential information using RNN.

\vspace{2mm}
\textbf{CNN-GRU} \citep{Wang_2023} combines CNN for spatial feature extraction with GRU for temporal modeling, improving county-level winter wheat yield predictions.

\vspace{2mm}
\textbf{UNet} \citep{Chakraborty_2023} is employed for bloom mapping and crop load estimation, demonstrating a strong correlation between bloom density and almond yield across different seasons.

\vspace{2mm}
\textbf{1D-CNN-GRU} \citep{Khan_2024} integrates 1D-CNN for feature extraction with GRU for sequential modeling, optimizing county-level corn and soybean yield predictions through crop-type and location transfer learning strategies.

\vspace{2mm}
\textbf{UNet-ConvLSTM} \citep{Kamangir_2024} integrates UNet for extracting spatial correlations and ConvLSTM for modeling temporal dynamics in yield progression.

\vspace{2mm}
\textbf{CTANet} \citep {Qu_2024} leverages attention mechanisms to effectively capture intricate relationships between spatial features and temporal patterns, enhancing crop yield prediction accuracy.

\vspace{2mm}
\textbf{SFC-DenseNet-AM} \citep{Seong_2024} effectively captures seasonal variations in crop yield while addressing the issues of imbalanced data by generating patches based on crop ratios.

\subsubsection{Evaluation measures}
We employ various primitive metrics to assess the performance of various methods including our proposed approach such as Mean Absolute Percentage Error (MAPE), Root Mean Squared Logarithmic Error (RMSLE), and Symmetric Mean Absolute Percentage Error (SMAPE). MAPE determines the average percentage difference between predicted and actual values, as shown in Eq.~(\ref{eq:mape}). RMSLE calculates the square root of mean squared logarithmic differences as presented in Eq.~(\ref{eq:rmsle}). SMAPE normalizes errors by using the average of actual with predicted values to ensure balanced evaluation in variable datasets, as shown in Eq.~(\ref{eq:smape}). 

\begin{equation}
\mathit{MAPE} = \frac{1}{N} \sum_{i=1}^{N} \left| \frac{Y_i - Y_{pred}}{Y_i} \right| \times 100    \tag{21}
\label{eq:mape}
\end{equation}

\begin{equation}
\mathit{RMSLE} = \sqrt{\frac{1}{N} \sum_{i=1}^{N} \left( \log(1 + Y_i) - \log(1 + Y_{pred}) \right)^2}      \tag{22}
\label{eq:rmsle}
\end{equation}

\begin{equation}
\mathit{SMAPE} = \frac{1}{N} \sum_{i=1}^{N} \frac{\left| Y_i - Y_{pred} \right|}{\frac{\left| Y_i \right| + \left| Y_{pred} \right|}{2}} \times 100      \tag{23}
\label{eq:smape}
\end{equation}
where, \( Y_i \) represents the actual value of the \(i\)-th sample, \( Y_{pred} \) denotes the predicted value of the \(i\)-th sample, and \( N \) is the total number of samples.

\subsection{Experimental results}
In this section, we present an overview of the experimental findings and performance comparisons for our proposed approach.
At the start, we assess how our method performs in relation to recent studies. We benchmark it against leading models in the field and explore the computational complexity. We conduct ablation studies to examine the impact and significance of each component within our framework.

\subsubsection{Comparative analysis with state-of-the-art literature}
\textbf{On Sentinel-1 dataset:}
Table~\ref{tab: Method(S1)} highlights the excellent performance of MTMS-YieldNet on the Sentinel-1 dataset. 
 MTMS-YieldNet outperforms other state-of-the-art methods by maintaining low error rates in the range of 1\%-3\%, making it highly effective for agricultural yield prediction. This performance of our method is attributed to learn patterns from the optical and radar time-series data. This provides a more comprehensive view of the crop growth cycle.
\begin{table*}[!ht] \footnotesize
\caption{Quantitative evaluation of the proposed MTMS-YieldNet with state-of-the-art yield prediction methods on the Sentinel-1 dataset. Lower MAPE, RMSLE, and SMAPE values indicate better performance. The best results are highlighted in \textcolor{darkgreen}{green}, the second-best in \textcolor{cyan}{cyan}, and the absolute performance drop relative to MTMS-YieldNet is represented by $|\nabla|$.}
\begin{tabular*}{\linewidth}{@{\extracolsep{\fill}}l ccc}
    \toprule
    \textbf{Model} & \textbf{MAPE$_{|\nabla|}$} & \textbf{RMSLE$_{|\nabla|}$} & \textbf{SMAPE$_{|\nabla|}$
    } \\
    \midrule
    GNN-RNN \citep{Fan_2022} & 0.365$_{0.029}$ & 0.796$_{0.299}$ & 0.512$_{0.150}$ \\
    UNet \citep{Chakraborty_2023} & 0.412$_{0.076}$ & 0.704$_{0.207}$ & 0.554$_{0.192}$ \\
    CNN-GRU \citep{Wang_2023} & 0.408$_{0.072}$ & 0.825$_{0.328}$ & 0.593$_{0.231}$ \\
    1D-CNN \citep{Khan_2024} & 0.400$_{0.064}$ & 0.790$_{0.293}$ & 0.572$_{0.210}$ \\
    UNet-ConvLSTM \citep{Kamangir_2024} & 0.383$_{0.047}$ & 0.792$_{0.295}$ & 0.598$_{0.236}$ \\
    CTANet \citep{Qu_2024} & 0.404$_{0.068}$ & 0.743$_{0.246}$ & 0.558$_{0.196}$ \\
    SFC-DenseNet-AM \citep{Seong_2024} & \textcolor{cyan}{0.346$_{0.010}$} & \textcolor{darkgreen}{0.449$_{0.048}$} & \textcolor{cyan}{0.409$_{0.047}$} \\
    \hline
    MTMS-YieldNet (Our proposed) & \textcolor{darkgreen}{0.336$_{0.000}$} & \textcolor{cyan}{0.497$_{0.000}$} & \textcolor{darkgreen}{0.362$_{0.000}$} \\
    \hline
\end{tabular*}
\label{tab: Method(S1)}
\end{table*}

\textbf{On Sentinel-2 dataset:}
Table~\ref{tab: Method(S2)} illustrates the results of MTMS-YieldNet on the Sentinel-2 dataset. The comparison demonstrates that methods such as those by \cite{Wang_2023} and \cite{Khan_2024} show accuracy drops of 4.2\%-6.7\% compared to MTMS-YieldNet. MTMS-YieldNet achieves better results with a more stable error margin of 0.3\%-2.5\% compared to other methods. This improvement results from MTMS-YieldNet's integration of temporal multi-scale features and to captures spatial and temporal dependencies effectively. In the evaluation, the model performs better in various environmental conditions and provides an enhancement in precision with reduced errors in different geographical areas.

\begin{table*}[!ht] \footnotesize
\caption{Quantitative evaluation of the proposed MTMS-YieldNet with state-of-the-art yield prediction methods on the Sentinel-2 dataset. Lower MAPE, RMSLE, and SMAPE values indicate better performance. The best results are highlighted in \textcolor{darkgreen}{green}, the second-best in \textcolor{cyan}{cyan}, and the absolute performance drop relative to MTMS-YieldNet is represented by $|\nabla|$.}

\begin{tabular*}{\linewidth}{@{\extracolsep{\fill}}l ccc}
    \toprule
    \textbf{Model} & \textbf{MAPE$_{|\nabla|}$} & \textbf{RMSLE$_{|\nabla|}$} & \textbf{SMAPE$_{|\nabla|}$} \\
    \midrule
    GNN-RNN \citep{Fan_2022} & 0.379$_{0.048}$ & 0.803$_{0.214}$ & 0.546$_{0.113}$ \\
    UNet \citep{Chakraborty_2023} & 0.436$_{0.105}$ & 0.754$_{0.165}$ & 0.635$_{0.202}$ \\
    CNN-GRU \citep{Wang_2023} & 0.421$_{0.090}$ & 0.924$_{0.335}$ & 0.624$_{0.191}$ \\
    1D-CNN \citep{Khan_2024} & 0.432$_{0.101}$ & 0.828$_{0.239}$ & 0.616$_{0.183}$ \\
    UNet-ConvLSTM \citep{Kamangir_2024} & 0.398$_{0.067}$ & 0.869$_{0.280}$ & 0.583$_{0.150}$ \\
    CTANet \citep{Qu_2024} & 0.401$_{0.070}$ & 0.795$_{0.206}$ & 0.575$_{0.142}$ \\
    SFC-DenseNet-AM \citep{Seong_2024} & \textcolor{cyan}{0.365$_{0.034}$} & \textcolor{cyan}{0.612$_{0.023}$} & \textcolor{cyan}{0.478$_{0.045}$} \\
    \hline
    MTMS-YieldNet (Our proposed) & \textcolor{darkgreen}{0.331$_{0.000}$} & \textcolor{darkgreen}{0.589$_{0.000}$} & \textcolor{darkgreen}{0.433$_{0.000}$}\\
    \hline
\end{tabular*}
\label{tab: Method(S2)}
\end{table*}

\textbf{On Landsat-8 dataset:}
Table~\ref{tab: Method(L8)} presents the results of MTMS-YieldNet on the Landsat-8 dataset by showing its superior performance compared to existing methods like \cite{Fan_2022} and \cite{Chakraborty_2023} with accuracy reductions ranging from 1.5\% to 3.2\%. 

\begin{figure}[h!]
    \centering
    \begin{minipage}{0.325\textwidth}
        \centering
        \includegraphics[width=\textwidth]{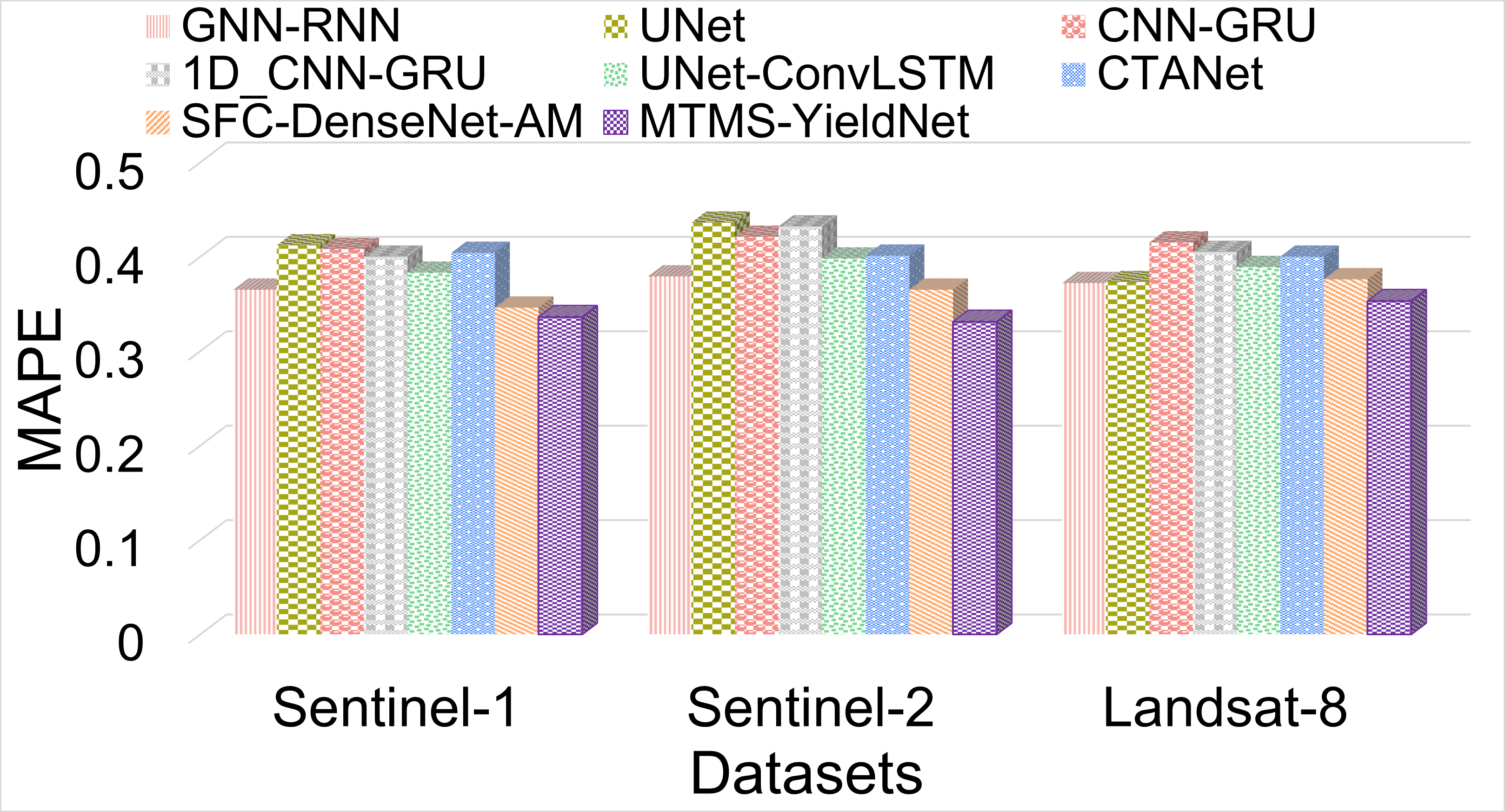}
        \caption*{(a) MAPE}
    \end{minipage}%
    \hspace{0.0125cm}
    \begin{minipage}{0.325\textwidth}
        \centering
        \includegraphics[width=\textwidth]{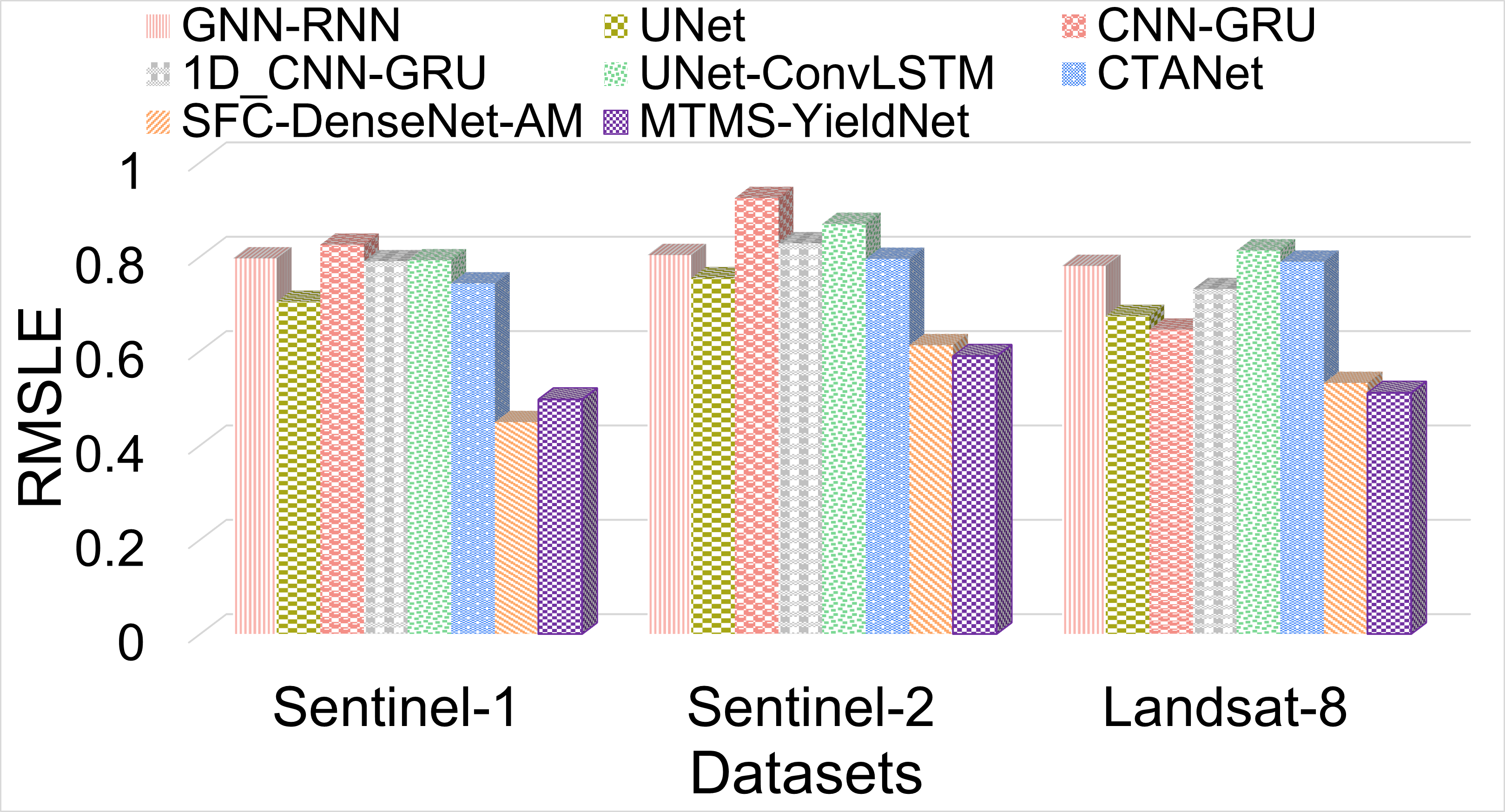}
        \caption*{(b) RMSLE}
    \end{minipage}%
    \hspace{0.0125cm}
    \begin{minipage}{0.325\textwidth}
        \centering
        \includegraphics[width=\textwidth]{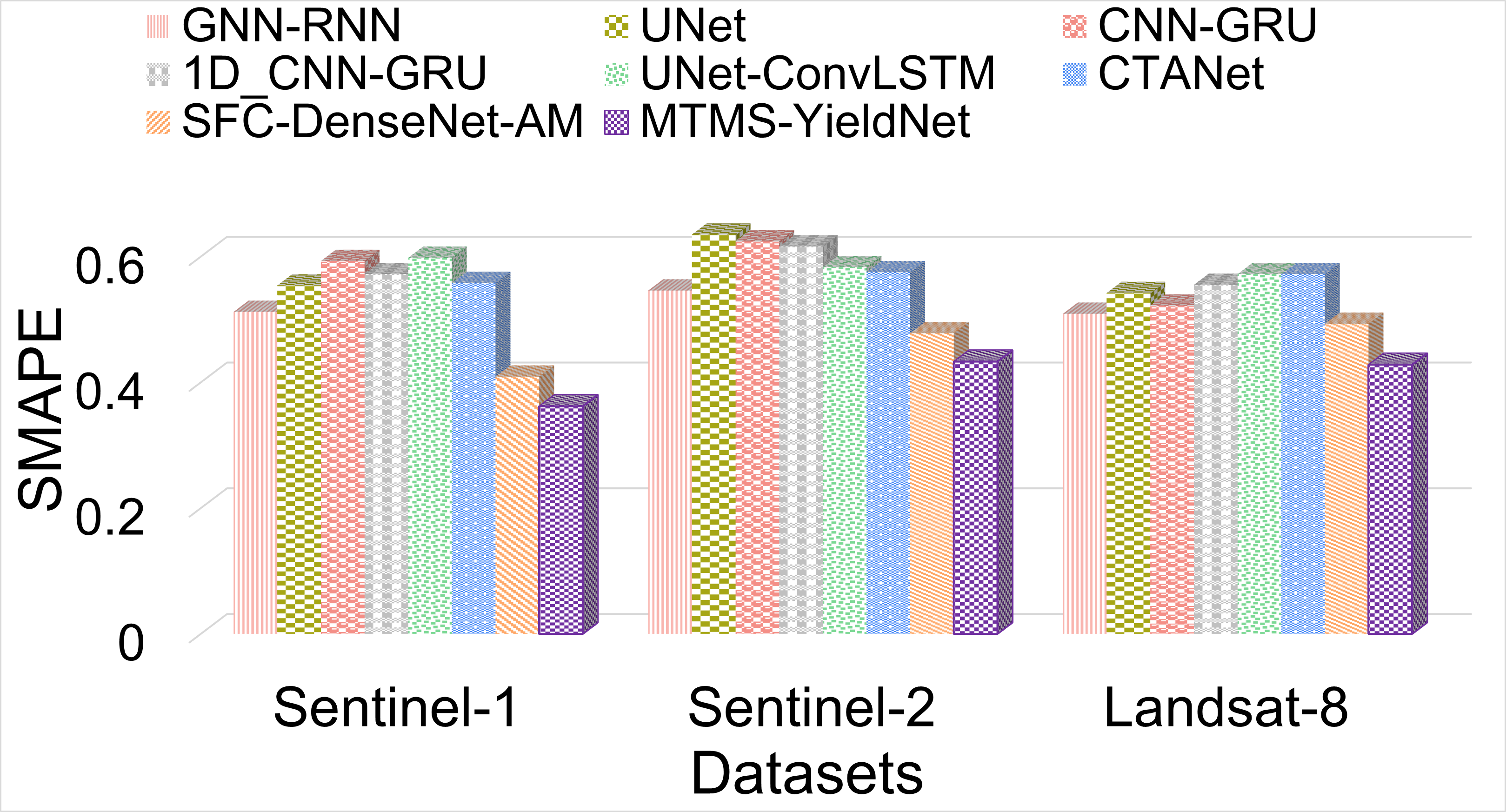}
        \caption*{(c) SMAPE}
    \end{minipage}
    \caption{Comparison of prediction performance using Sentinel-1, Sentinel-2, and Landsat-8 datasets: (a) MAPE, (b) RMSLE, and (c) SMAPE, with paddy yield measured per acre.}
  \label{fig: pfc}
\end{figure}

\begin{table*}[!ht] \footnotesize
\caption{Quantitative evaluation of the proposed MTMS-YieldNet with state-of-the-art yield prediction methods on the Landsat-8 dataset. Lower MAPE, RMSLE, and SMAPE values indicate better performance. The best results are highlighted in \textcolor{darkgreen}{green}, the second-best in \textcolor{cyan}{cyan}, and the absolute performance drop relative to MTMS-YieldNet is represented by $|\nabla|$.}
\begin{tabular*}{\linewidth}{@{\extracolsep{\fill}}l ccc}
    \toprule
    \textbf{Model} & \textbf{MAPE$_{|\nabla|}$} & \textbf{RMSLE$_{|\nabla|}$} & \textbf{SMAPE$_{|\nabla|}$
    } \\
    \midrule
    GNN-RNN \citep{Fan_2022} & \textcolor{cyan}{0.372$_{0.019}$} & 0.780$_{0.269}$ & 0.509$_{0.081}$ \\
    UNet \citep{Chakraborty_2023} & 0.373$_{0.020}$ & 0.674$_{0.163}$ & 0.542$_{0.114}$ \\
    CNN-GRU \citep{Wang_2023} & 0.415$_{0.062}$ & 0.645$_{0.134}$ & 0.522$_{0.094}$ \\
    1D-CNN \citep{Khan_2024} & 0.405$_{0.052}$ & 0.731$_{0.220}$ & 0.555$_{0.127}$ \\
    UNet-ConvLSTM \citep{Kamangir_2024} & 0.389$_{0.036}$ & 0.813$_{0.302}$ & 0.573$_{0.145}$ \\
    CTANet \citep{Qu_2024} & 0.400$_{0.047}$ & 0.789$_{0.278}$ & 0.572$_{0.144}$ \\
   SFC-DenseNet-AM \citep{Seong_2024} & 0.376$_{0.023}$ & \textcolor{cyan}{0.532$_{0.021}$} & \textcolor{cyan}{0.493$_{0.065}$} \\
    \hline
    MTMS-YieldNet (Our proposed) & \textcolor{darkgreen}{0.353$_{0.000}$} & \textcolor{darkgreen}{0.511$_{0.000}$} & \textcolor{darkgreen}{0.428$_{0.000}$} \\
    \hline
\end{tabular*}
\label{tab: Method(L8)}
\end{table*}
The model also handles class imbalance more effectively with a smaller error margin (2\% to 4\%) than other techniques. 
MTMS-YieldNet performs well because of its ability to integrate spatial and temporal features across multiple scales and capture both long-term trends and short-term variations in crop growth. Its advanced feature learning capabilities help the model detect subtle changes in vegetation and ensure accurate yield predictions throughout different seasons.
Fig.~\ref{fig: pfc} performs the comparative analysis between the datasets Sentinel-1, Sentinel-2, and Landsat-8. Our proposed model MTMS-YieldNet achieves better performance in comparison to seven state-of-the-art methods, including GNN-RNN, CNN-GRU, UNet, 1D-CNN, UNet-ConvLSTM, CTANet, and SFC-DenseNet-AM, for predicting crop yield one year ahead.

\subsubsection{Comparative analysis with state-of-the-art models}
In this section, we conduct a thorough quantitative comparison of our proposed framework with SOTA models to assess its effectiveness. We used the SICKLE dataset \citep{Sani_2024}, which includes multi-sensor satellite imagery from Sentinel-1 (S1), Sentinel-2 (S2), and Landsat-8 (L8).

\begin{table*}[!ht] \footnotesize
\caption{Quantitative evaluation of the proposed MTMS-YieldNet with state-of-the-art (SOTA) models on the Sentinel-1, Sentinel-2, and Landsat-8 datasets. Lower MAPE, RMSLE, and SMAPE values indicate better performance. The best results are highlighted in \textcolor{darkgreen}{green}, the second-best in \textcolor{cyan}{cyan}, and the absolute performance drop relative to MTMS-YieldNet is represented by $|\nabla|$.}
\begin{tabular*}{\linewidth}{@{\extracolsep{\fill}}l l cccc}
    \toprule
    \textbf{Dataset} & \textbf{Model} & \textbf{MAPE$_{|\nabla|}$} & \textbf{RMSLE$_{|\nabla|}$} & \textbf{SMAPE$_{|\nabla|}$} \\
    \midrule
     \addlinespace
    \multirow{7}{*}{$Sentinel-1$} 
    & ResNet152 \citep{He_2016} &  0.440$_{0.104}$ & 0.674$_{0.177}$ & 0.550$_{0.188}$  \\ 
    & DenseNet121 \citep{huang2017densely} &  0.386$_{0.050}$ & 0.731$_{0.234}$ & 0.537$_{0.175}$   \\ 
    & BiLSTM  \citep{Ghasemlounia_2021} & 0.391$_{0.055}$ & 0.720$_{0.223}$ & 0.547$_{0.185}$   \\ 
    & StackedLSTM \citep{Maaliw_2021} &  \textcolor{cyan}{0.350$_{0.014}$} & \textcolor{cyan}{0.663$_{0.166}$} & \textcolor{cyan}{0.466$_{0.104}$} \\
    & ResNet-LSTM \citep{Choi_2018} &  0.418$_{0.082}$ & 0.868$_{0.371}$ & 0.659$_{0.297}$  \\
    & CNN-LSTM  \citep{Islam_2020}  &  0.381$_{0.045}$ & 0.693$_{0.196}$ & 0.481$_{0.119}$ \\
    & CLIP(ViT-B/32) \citep{pmlr-v139-radford21a} &  0.400$_{0.064}$ & 0.790$_{0.293}$ & 0.572$_{0.210}$  \\
    \midrule
    & MTMS-YieldNet (Our proposed) & \textcolor{darkgreen}{0.336$_{0.000}$} & \textcolor{darkgreen}{0.497$_{0.000}$} & \textcolor{darkgreen}{0.362$_{0.000}$} \\
    \hline
    \addlinespace
    \multirow{7}{*}{$Sentinel-2$} 
    & Renset152 \citep{He_2016} & 0.428$_{0.097}$ & \textcolor{cyan}{0.668$_{0.079}$} & 0.535$_{0.102}$  \\ 
    & Denset121 \citep{huang2017densely} & 0.379$_{0.048}$ & 0.803$_{0.214}$ & 0.546$_{0.113}$  \\ 
    & BiLSTM  \citep{Ghasemlounia_2021}  & 0.383$_{0.052}$ & 0.930$_{0.341}$ & 0.531$_{0.098}$  \\ 
    & StackedLSTM \citep{Maaliw_2021} & 0.459$_{0.128}$ & 0.899$_{0.310}$ & 0.599$_{0.166}$  \\
    & Resnet-LSTM  \citep{Choi_2018}& 0.396$_{0.065}$ & 0.946$_{0.357}$ & 0.634$_{0.201}$  \\
    & CNN-LSTM  \citep{Islam_2020}  & \textcolor{cyan}{0.376$_{0.045}$} & 0.698$_{0.109}$ & \textcolor{cyan}{0.500$_{0.067}$}  \\
   &  CLIP(ViT-B/32) \citep{pmlr-v139-radford21a} & 0.389$_{0.058}$ & 0.749$_{0.160}$ & 0.545$_{0.112}$  \\
    \midrule
    & MTMS-YieldNet(Our proposed) & \textcolor{darkgreen}{0.331$_{0.000}$} & \textcolor{darkgreen}{0.589$_{0.000}$} & \textcolor{darkgreen}{0.433$_{0.000}$} \\
    \hline
    \addlinespace
     \multirow{7}{*}{$Landsat-8$} 
    & Renset152 \citep{He_2016} &  0.439$_{0.086}$ & 0.657$_{0.146}$ & 0.537$_{0.109}$  \\ 
    & Denset121 \citep{huang2017densely} &  0.389$_{0.036}$ & 0.713$_{0.202}$ & 0.554$_{0.126}$  \\ 
    & BiLSTM  \citep{Ghasemlounia_2021}  &  \textcolor{cyan}{0.373$_{0.020}$} & \textcolor{cyan}{0.654$_{0.143}$} & \textcolor{cyan}{0.489$_{0.061}$} \\ 
    & StackedLSTM \citep{Maaliw_2021} &  0.376$_{0.023}$ & 0.659$_{0.148}$ & 0.496$_{0.068}$ \\
    & Resnet-LSTM  \citep{Choi_2018}&   0.427$_{0.074}$ & 0.779$_{0.268}$ & 0.605$_{0.177}$ \\
    & CNN-LSTM  \citep{Islam_2020}  &   0.397$_{0.044}$ & 0.679$_{0.168}$ & 0.531$_{0.103}$ \\
    & CLIP(ViT-B/32) \citep{pmlr-v139-radford21a} &  0.403$_{0.050}$ & 0.751$_{0.240}$ & 0.561$_{0.133}$ \\
    \midrule
    & MTMS-YieldNet(Our proposed) &  \textcolor{darkgreen}{0.353$_{0.000}$} & \textcolor{darkgreen}{0.511$_{0.000}$} & \textcolor{darkgreen}{0.428$_{0.000}$}\\
    \hline
\end{tabular*}
\label{tab: Model}
\vspace{0.8mm}
\end{table*}
Our framework was compared with a diverse set of SOTA models relevant to agricultural and environmental data analysis. The models used for comparison include ResNet152 \citep{He_2016}, DenseNet121 \citep{huang2017densely}, BiLSTM \citep{Ghasemlounia_2021}, StackedLSTM \citep{Maaliw_2021}, ResNet-LSTM \citep{Choi_2018}, CNN-LSTM \citep{Islam_2020}, and CLIP (ViT-B/32) \citep{pmlr-v139-radford21a}.  

Table~\ref{tab: Model} presents the comparative analysis of our proposed method's performance with other SOTA models for crop yield prediction. MTMS-YieldNet consistently delivers better results across Sentinel-2 (S2), Sentinel-1 (S1), and Landsat-8 (L8) datasets, achieving the lowest values for MAPE, RMSLE, and SMAPE. Notably, MTMS-YieldNet surpasses StackedLSTM \citep{Maaliw_2021} on S1 by 3.75\% in MAPE and outperforms Densenet121 \citep{huang2017densely} on S2 and L8 by 4.76\% and 3.64\% in MAPE, respectively. CLIP(ViT-B/32) \citep{pmlr-v139-radford21a} demonstrates competitive performance but lags behind MTMS-YieldNet across all metrics.

\subsubsection{Ablation analysis}
To evaluate the impact of each key module in the MTMS-YieldNet framework, we conducted an ablation study by individually removing each module and observing the effects on the S2 dataset from SICKLE \citep{Sani_2024}. Our results indicate that each proposed component and strategy significantly enhances the overall performance of our framework. We specifically examined the roles of Multi-Spectral and Multi-Temporal Image Preprocessing, Data Augmentation, Spatio-Temporal Feature Extraction, and Feature Subset Selection. Section \ref{sec: method} gives a detailed overview of the proposed model pipeline.

\begin{table*}[!ht] \footnotesize
\caption{Results for image preprocessing models.}
\begin{tabular*}{\linewidth}{@{\extracolsep{\fill}}l ccc}
    \toprule
    \multicolumn{4}{c}{\textbf{Image Preprocessing}} \\
    \cmidrule(l){1-4}
    \textbf{Model} & \textbf{MAPE$_{|\nabla|}$} & \textbf{RMSLE$_{|\nabla|}$} & \textbf{SMAPE$_{|\nabla|}$} \\
    \midrule
    MTMS-YieldNet(No Enhancement) & 0.395$_{0.064}$ & 0.823$_{0.234}$ & 0.569$_{0.136}$ \\
    MTMS-YieldNet(SIFT) & 0.362$_{0.031}$ & 0.612$_{0.023}$ & 0.452$_{0.019}$ \\
    MTMS-YieldNet(NR) & 0.375$_{0.044}$ & 0.601$_{0.012}$ & 0.457$_{0.024}$ \\
    MTMS-YieldNet(AC) & 0.368$_{0.037}$ & 0.758$_{0.169}$ & 0.508$_{0.075}$ \\
    \hline
    \textbf{MTMS-YieldNet(Laplacian)} & \textbf{0.331$_{0.000}$} & \textbf{0.589$_{0.000}$} & \textbf{0.433$_{0.000}$} \\
    \hline
\end{tabular*}
\label{tab: Preprocessing}
\vspace{0.8mm}
\end{table*}

\textit{Effect of different preprocessing techniques:} The performance of MTMS-YieldNet shows substantial improvements across various preprocessing techniques, as summarized in Table~\ref{tab: Preprocessing}. 
The SIFT technique improves MAPE by 6.5\%, but not as effectively as the Laplacian filter. When we apply noise reduction (NR)
, the model shows a minor improvement in RMSLE by 3.6\% but an increase in MAPE by 4.4\%, indicating that while noise reduction can refine data, it does not necessarily improve the prediction model in this context. This might be because noise reduction can sometimes remove important details along with the noise. Atmospheric correction (AR), though targeting environmental influences, results in a slight increase in MAPE by 2.6\% and an increase in SMAPE by 5.4\%, indicating that it may not offer a clear benefit over other techniques for crop yield prediction. This is likely because atmospheric correction can introduce artifacts that hinder model performance. However, Laplacian leads to the most substantial reduction in MAPE by 15.5\%, outperforming other preprocessing methods. This approach improves feature extraction by enhancing the visibility of critical features while mitigating noise and contrast issues in the dataset. 

\begin{figure*}[!ht]
    \begin{minipage}{0.7\textwidth}  
        \footnotesize
        \captionof{table}{Ablation study for synthetic image generation models for augmentation.}
        \centering
        \renewcommand{\arraystretch}{1.4}
        \resizebox{0.95\textwidth}{!}{  
        \begin{tabular}{l ccc}
            \toprule
            \multicolumn{4}{c}{\textbf{Synthetic Image Generation}} \\
            \cmidrule(l){1-4}
            \textbf{Model} & \textbf{MAPE$_{|\nabla|}$} & \textbf{RMSLE$_{|\nabla|}$} & \textbf{SMAPE$_{|\nabla|}$} \\
            \midrule
            MTMS-YieldNet (Traditional) & 0.365$_{0.034}$ & 0.612$_{0.023}$ & 0.478$_{0.045}$ \\
            MTMS-YieldNet (DC GAN) & 0.347$_{0.016}$ & 0.626$_{0.037}$ & 0.464$_{0.031}$ \\
            MTMS-YieldNet (Vanilla GAN) & 0.359$_{0.028}$ & 0.619$_{0.030}$ & 0.468$_{0.035}$ \\
            \hline
            \textbf{MTMS-YieldNet (Diffusion Model)} & 
            \textbf{0.331$_{0.000}$} & \textbf{0.589$_{0.000}$} & \textbf{0.433$_{0.000}$} \\
            \hline
        \end{tabular}}
        \label{tab: Synthetic}
        \vspace{0.8mm}
    \end{minipage}%
    \hfill
    \begin{minipage}{0.01\textwidth}  
        \centering
        \begin{tikzpicture}
            \draw[dashed] (0,0) -- (0,3.8);  
        \end{tikzpicture}
    \end{minipage}%
    \hfill
    \begin{minipage}{0.25\textwidth}  
        \centering
        \includegraphics[width=\linewidth]{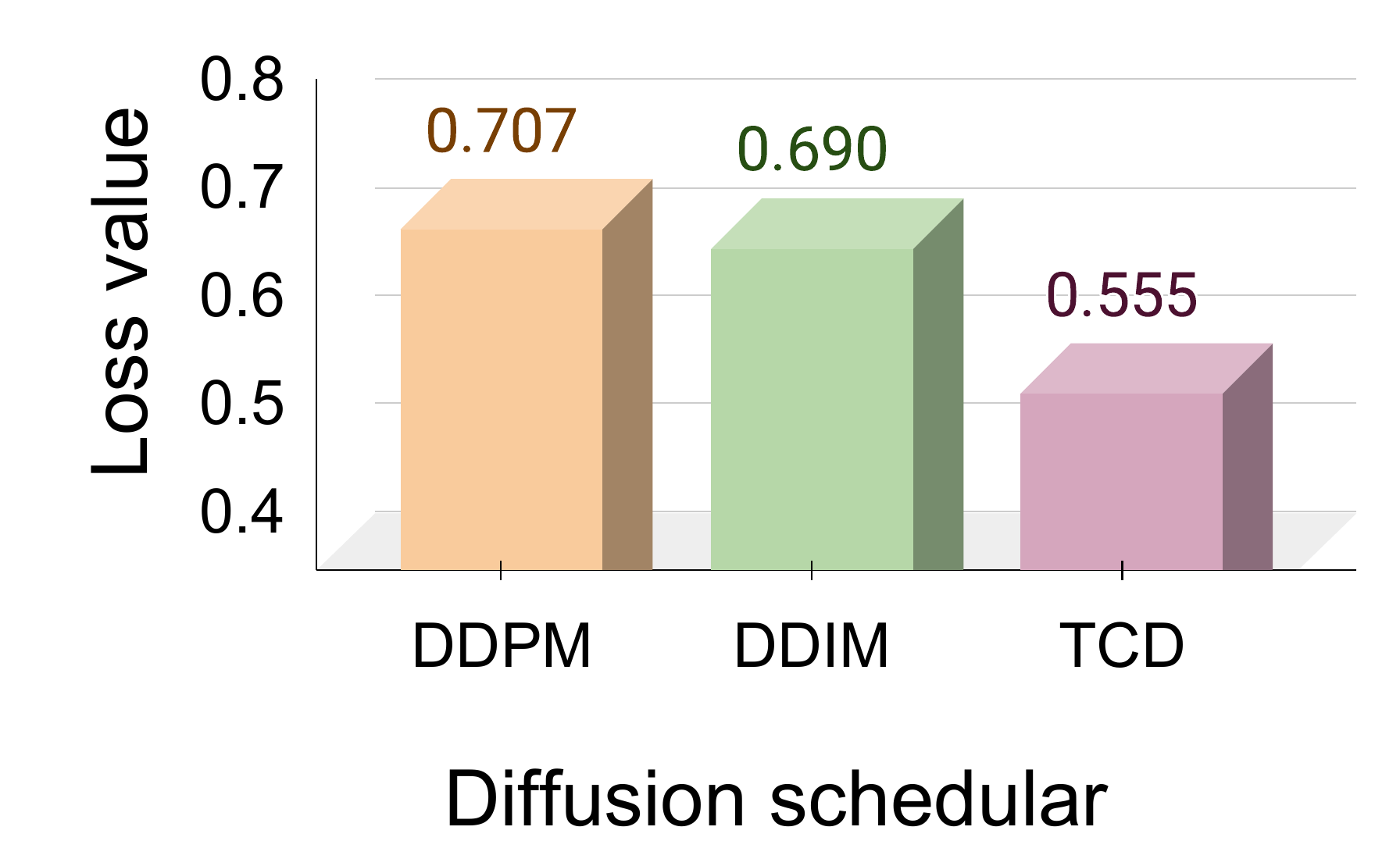}
        \caption{Comparison of diffusion noise schedulers.}
        \label{fig: diffusion_loss}
    \end{minipage}
\end{figure*}

\textit{Synthetic Image Generation \& Diffusion Noise Schedulers:} Synthetic image generation analysis in Table~\ref{tab: Synthetic} reveals that employing traditional augmentation techniques, such as rotation and flipping, results in an improvement in MAPE by 5.2\% compared to models without augmentation. However, the application of DC-GAN \citep{Ma_2020} results in even greater performance improvements, yielding a 5.2\% reduction in MAPE and a 2.4\% improvement in SMAPE.
On the other hand, Vanilla GAN achieves improvements in MAPE and RMSLE but still falls short of the performance gains achieved by DC-GAN. Notably, the use of the Diffusion Model with TCD Scheduler \citep{zheng2024trajectory} yields the most substantial performance gains, reducing MAPE by 9.3\%, RMSLE by 3.9\%, and SMAPE by 7.8\%, compared to models using traditional augmentation methods such as rotation and flipping. These findings highlight the effectiveness of advanced diffusion-based models in image generation, providing superior performance in enhancing satellite imagery. 
The TCD Scheduler outperforms other schedulers like Denoising Diffusion Probabilistic Models (DDPMs) \citep{Ho2020} and Denoising Diffusion Implicit Models (DDIMs) \citep{Song2021} in diffusion-based image generation, showcasing its potential for improving model accuracy and generalization, as shown in Fig.~\ref{fig: diffusion_loss}. The model’s ability to generate high-quality synthetic samples while preserving critical features is pivotal for enhancing predictive performance in time-series satellite imagery tasks.

\begin{figure*}[ht]
 \centering
        \includegraphics[width=0.85\textwidth]{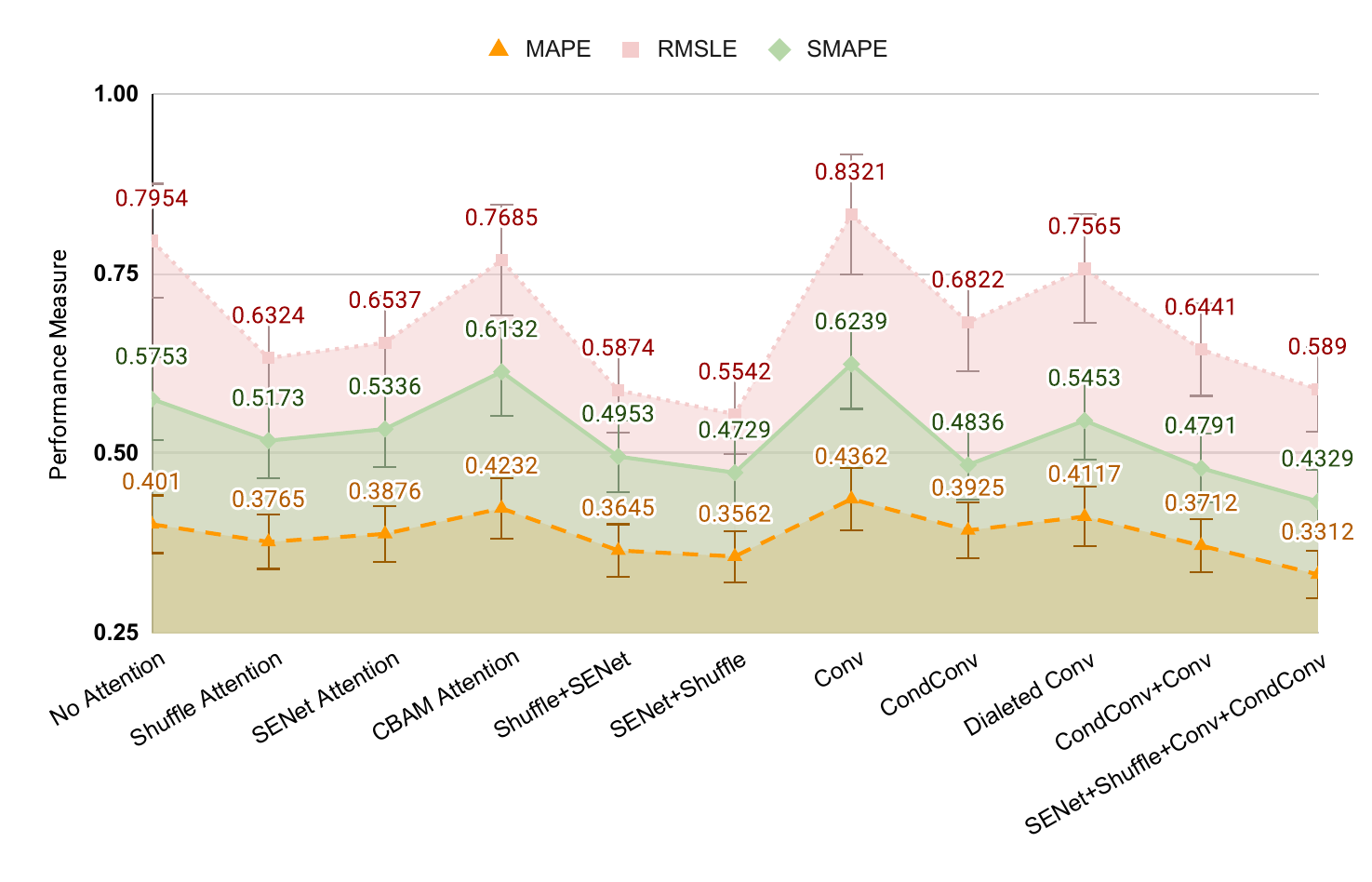}
        \caption{Error evaluation via comprehensive ablation study with attention mechanisms and convolution-based models.}
        \label{fig:graph1}
\end{figure*}

\begin{table*}[!ht] \footnotesize
\caption{Ablation study with different attention mechanisms and convolution-based models.}
\begin{tabular*}{\linewidth}{@{\extracolsep{\fill}}l ccc}
    \toprule
    \multicolumn{4}{c}{\textbf{Attention and Convolution Based Comparison}} \\
    \cmidrule(l){1-4}
    \textbf{Model} & \textbf{MAPE$_{|\nabla|}$} & \textbf{RMSLE$_{|\nabla|}$} & \textbf{SMAPE$_{|\nabla|}$} \\
    \midrule
     MTMS-YieldNet(No Attention) & 0.401$_{0.070}$ & 0.795$_{0.206}$ & 0.575$_{0.142}$ \\
     MTMS-YieldNet(Shuffle) & 0.377$_{0.046}$ & 0.632$_{0.043}$ & 0.517$_{0.084}$ \\
     MTMS-YieldNet(SENet) & 0.388$_{0.057}$ & 0.654$_{0.065}$ & 0.534$_{0.101}$ \\
     MTMS-YieldNet(CBAM) & 0.423$_{0.092}$ & 0.769$_{0.180}$ & 0.613$_{0.180}$ \\
     MTMS-YieldNet(Shuffle+SENet) & 0.365$_{0.034}$ & 0.587$_{0.002}$ & 0.495$_{0.062}$ \\
     MTMS-YieldNet(SENet+Shuffle) & 0.356$_{0.025}$ & 0.554$_{0.035}$ & 0.473$_{0.040}$ \\
     MTMS-YieldNet(Conv) & 0.436$_{0.105}$ & 0.832$_{0.243}$ & 0.624$_{0.191}$ \\
     MTMS-YieldNet(CondConv) & 0.393$_{0.062}$ & 0.682$_{0.093}$ & 0.484$_{0.051}$ \\
     MTMS-YieldNet(Dilated Conv) & 0.412$_{0.081}$ & 0.757$_{0.168}$ & 0.545$_{0.112}$ \\
     MTMS-YieldNet(CondConv+Conv) & 0.371$_{0.040}$ & 0.644$_{0.055}$ & 0.479$_{0.046}$ \\
    \hline
    \textbf{MTMS-YieldNet(SENet+Shuffle, Conv+CondConv)} & \textbf{0.331$_{0.000}$} & \textbf{0.589$_{0.000}$} & \textbf{0.433$_{0.000}$} \\
    \hline
\end{tabular*}
\label{tab: ATTENTION}
\vspace{0.8mm}
\end{table*}

\textit{Attention and convolution-based comparison:} The analysis of attention and convolution-based models, as shown in Table~\ref{tab: ATTENTION}, demonstrates how different combinations of attention mechanisms and convolution techniques impact model performance. When no attention mechanism is applied, using the shuffle technique results in a 6.1\% improvement in MAPE. 
The SENet 
and Dilated Convolution 
improve performance but still fall behind the attention-based models, with CondConv improving by 1.9\% and Dilated Convolution by 2.5\%. Combining convolution methods, such as Conv + CondConv, reduces the MAPE by 6.8\% but still doesn't match the best attention-based models.  Ultimately, combining shuffle + SENet with Conv + CondConv yields the best performance, resulting in a 17.4\% improvement in MAPE. Fig.~\ref{fig:graph1} highlights the effectiveness of hybrid models that integrate attention mechanisms and convolution techniques to improve model performance.  

\textit{Feature selection based study:} The analysis of optimizer-based models in Table~\ref{tab: optimizer} shows how different optimization techniques affect model performance. Using no optimizer serves as the baseline. The Golden
\begin{table*}[!ht] \footnotesize
\caption{Ablation results for optimizer-based models.}
\begin{tabular*}{\linewidth}{@{\extracolsep{\fill}}l ccc}
    \toprule
    \multicolumn{4}{c}{\textbf{Optimizer Based Comparisons}} \\
    \cmidrule(l){1-4}
    \textbf{Model} & \textbf{MAPE$_{|\nabla|}$} & \textbf{RMSLE$_{|\nabla|}$} & \textbf{SMAPE$_{|\nabla|}$} \\
    \midrule
     MTMS-YieldNet(No Optimizer) & 0.352$_{0.021}$ & 0.692$_{0.103}$ & 0.563$_{0.130}$\\
     MTMS-YieldNet(Golden Ratio) & 0.376$_{0.045}$ & 0.635$_{0.046}$ & 0.524$_{0.091}$ \\
     MTMS-YieldNet(Sail Fish) & 0.350$_{0.019}$ & 0.618$_{0.029}$ & 0.478$_{0.045}$ \\
    \hline
   \textbf{MTMS-YieldNet(Equilibrium Optimizer)} & \textbf{0.331$_{0.000}$} & \textbf{0.589$_{0.000}$} & \textbf{0.433$_{0.000}$} \\
    \hline
\end{tabular*}
\label{tab: optimizer}
\vspace{0.8mm}
\end{table*} 
 Ratio optimizer \citep{Nematollahi_2019} results in a 6.9\% increase in MAPE, likely because it may not effectively navigate the loss landscape for this problem. The Sail Fish optimizer \citep{Shadravan_2019} provides a smaller improvement of 0.5\%. The Equilibrium Optimizer \citep{Faramarzi_2020} yields the best performance with a 5.8\% reduction in MAPE. For RMSLE, the Golden Ratio optimizer shows an 8.3\% increase, while the Sail Fish optimizer achieves a 10.7\% improvement. The Equilibrium Optimizer leads with a 14.9\% reduction. Regarding SMAPE, the Golden Ratio optimizer improves by 7.0\%, the Sail Fish optimizer by 15.1\%, and the Equilibrium Optimizer shows the highest improvement of 23.1\%. This demonstrates that the Equilibrium Optimizer's balanced approach to exploration and exploitation makes it more suitable for this yield prediction task.

\section{Discussion}
\label{sec: Dis}
\subsection{Comparison of predicted vs actual yield}
\begin{figure}[h!]
    \centering
    \begin{minipage}{0.32\textwidth}
        \centering
        \includegraphics[width=\textwidth]{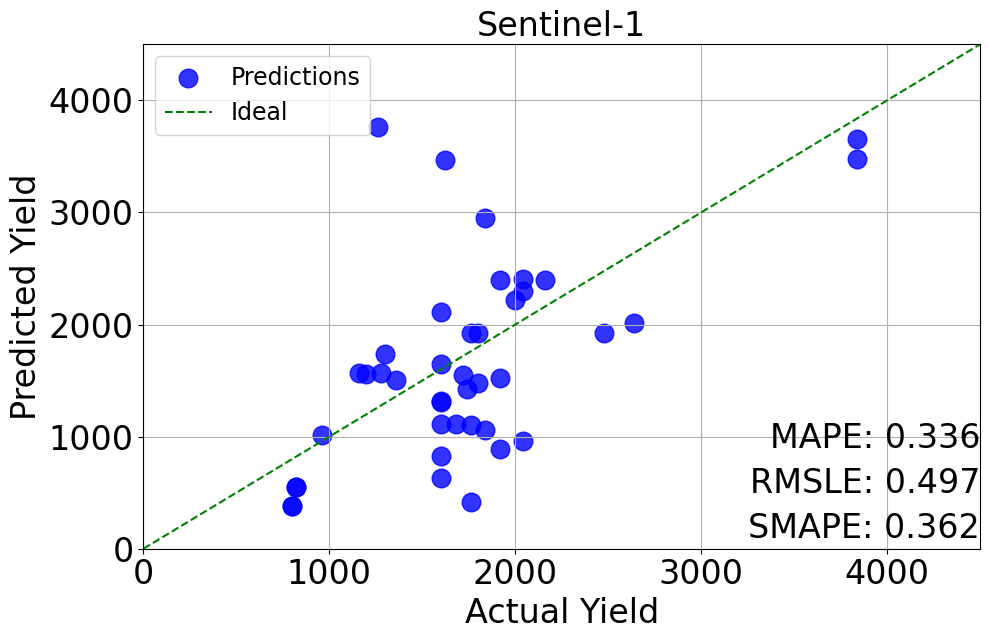}
    \end{minipage}%
     \hspace{0.2cm}
    \begin{minipage}{0.32\textwidth}
        \centering
        \includegraphics[width=\textwidth]{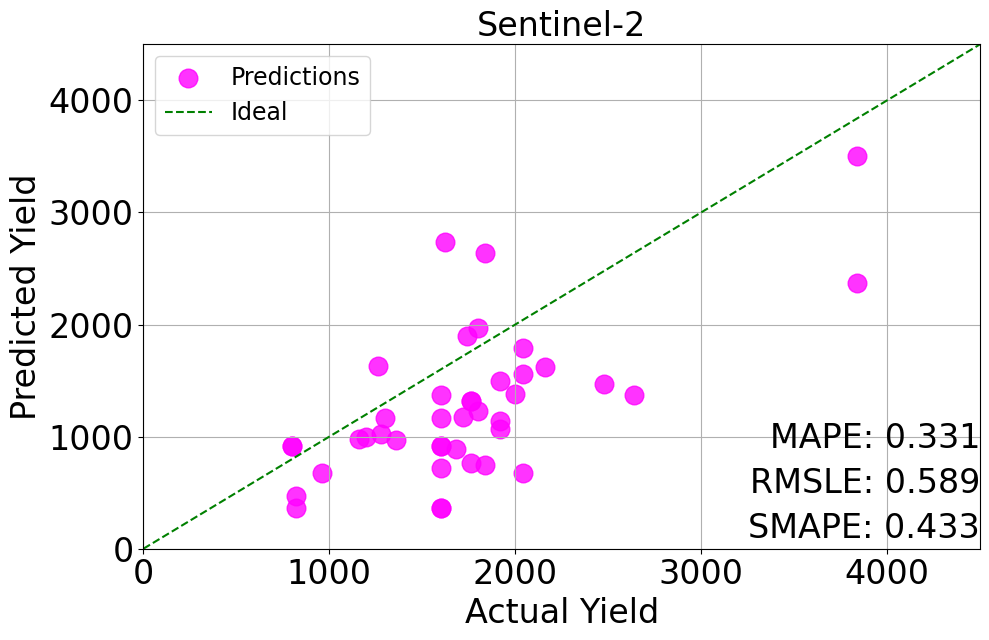}
    \end{minipage}%
     \hspace{0.2cm}
    \begin{minipage}{0.32\textwidth}
        \centering
        \includegraphics[width=\textwidth]{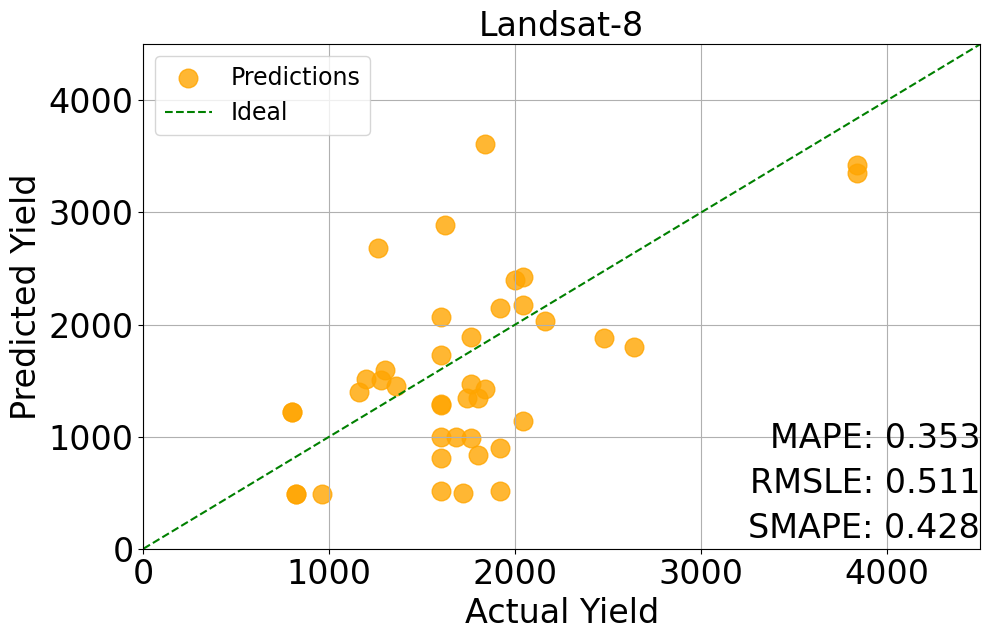}
    \end{minipage}
    \caption{Comparison of predicted vs actual yield over Sentinel-1, Sentinel-2 and Landsat-8 datasets.}
    \label{fig: yield}
\end{figure}

Based on the provided graph (see Fig.~\ref{fig: yield}), for the Landsat-8 dataset, the majority of predictions align closely with the ideal line, suggesting reasonable model performance. However, deviations are observed for certain data points, with metrics indicating a MAPE of 0.3530, RMSLE of 0.5109, and SMAPE of 0.4580. In contrast, the predictions for the Sentinel-1 dataset exhibit a closer alignment with the ideal line, reflecting better performance. The metrics for Sentinel-1 indicate a MAPE of 0.3364, RMSLE of 0.4966, and SMAPE of 0.3619, highlighting consistent predictions with a few deviations at higher yield values. Furthermore, for the Sentinel-2 dataset, the predictions show a slightly better alignment than Landsat-8 but with some deviations at higher yield values. The metrics for Sentinel-2 indicate a MAPE of 0.3312, RMSLE of 0.5890, and SMAPE of 0.4328.

\subsection{Comparison of metrics over the years}
\begin{figure}[h!]
    \centering
    \begin{minipage}{0.31\textwidth}
        \centering
        \includegraphics[width=\textwidth]{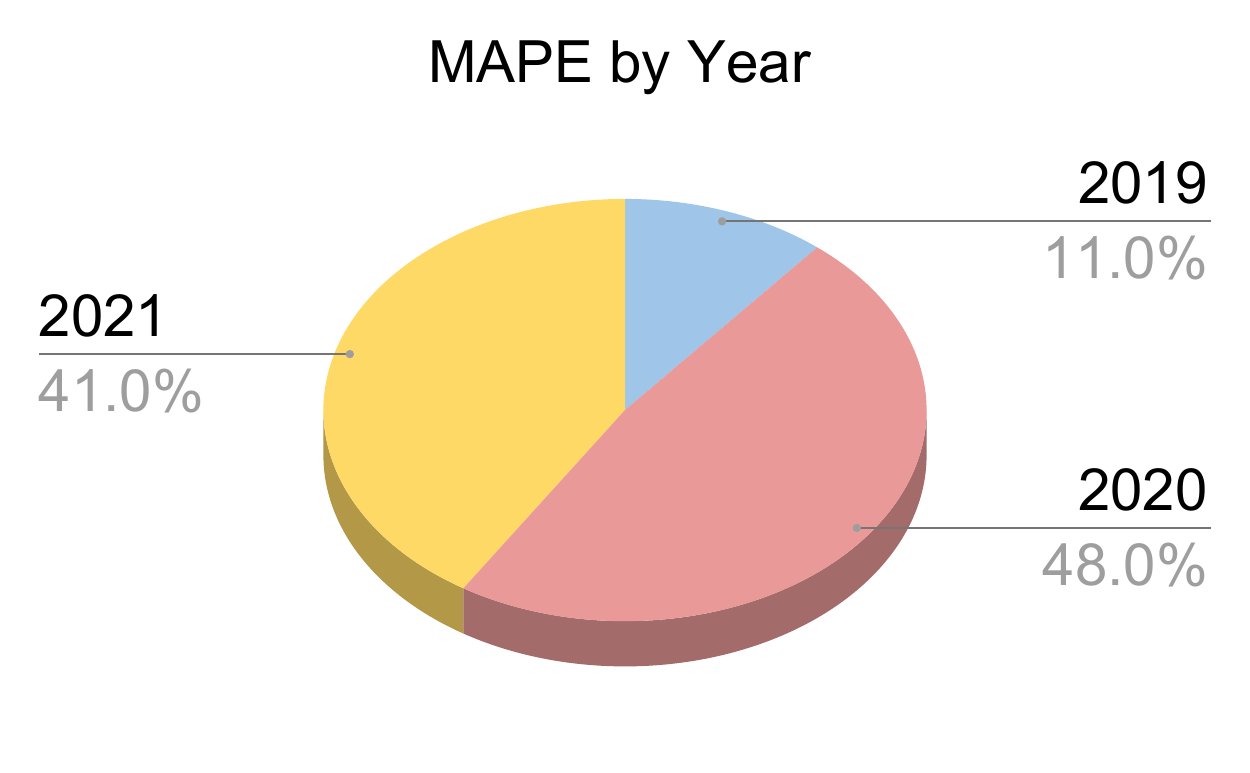}
        \caption*{(a) MAPE}
    \end{minipage}%
    \hspace{0.0125cm}
    \begin{minipage}{0.31\textwidth}
        \centering
        \includegraphics[width=\textwidth]{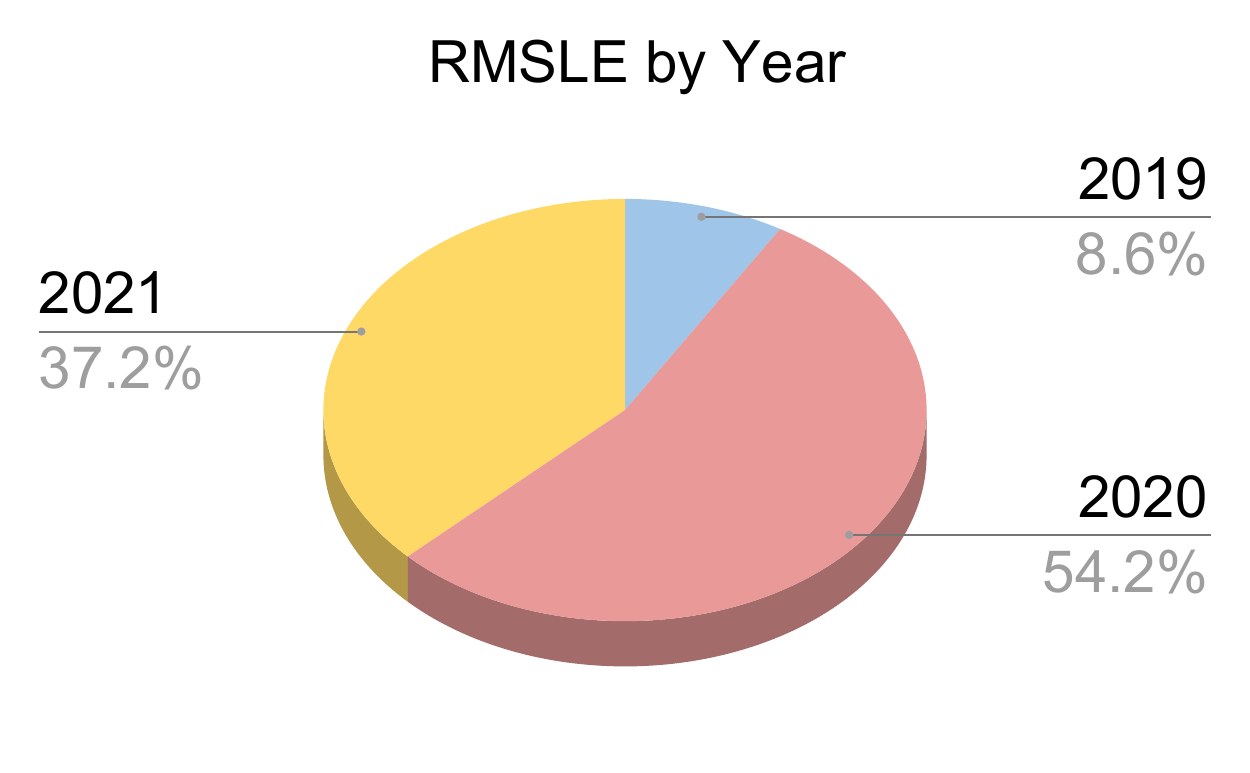}
        \caption*{(b) RMSLE}
    \end{minipage}%
    \hspace{0.0125cm}
    \begin{minipage}{0.31\textwidth}
        \centering
        \includegraphics[width=\textwidth]{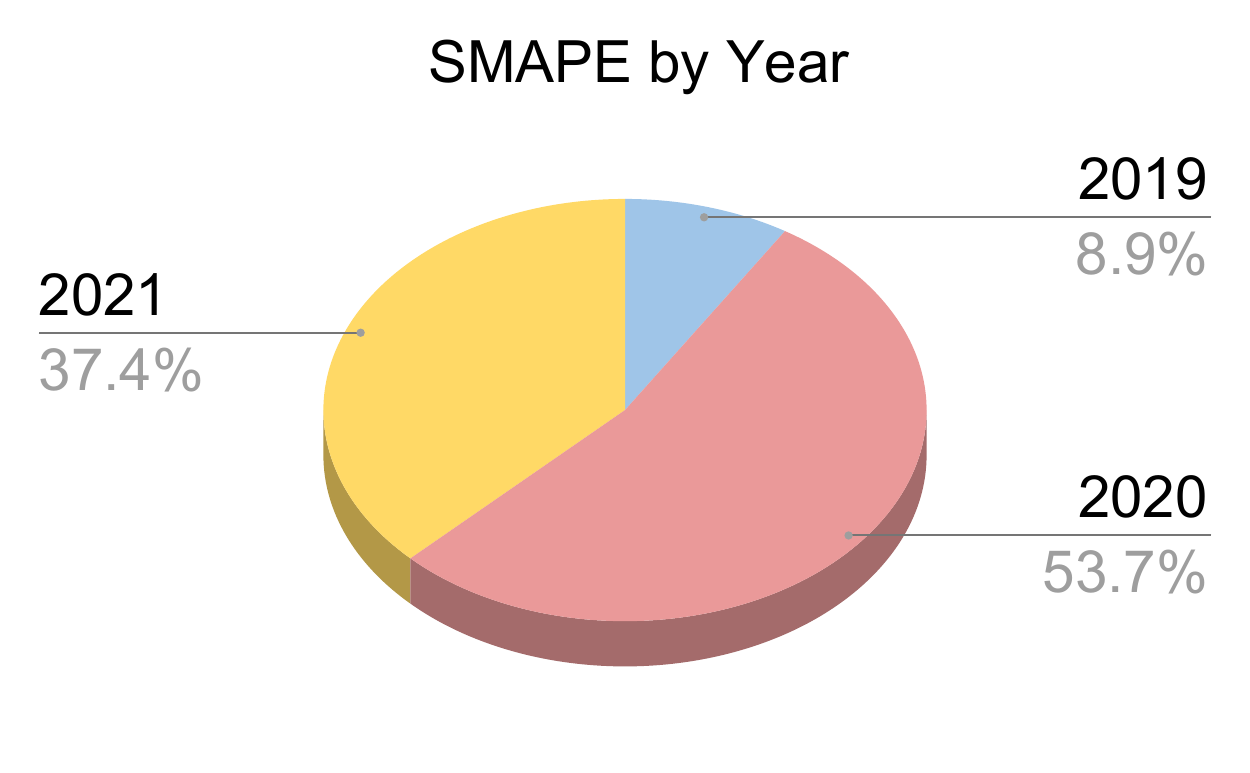}
        \caption*{(c) SMAPE}
    \end{minipage}
    \caption{ Comparison of metrics over the years for the Sentinel-1 dataset.}
  \label{fig: years_s1}
\end{figure}

\begin{figure}[h!]
    \centering
    \begin{minipage}{0.31\textwidth}
        \centering
        \includegraphics[width=\textwidth]{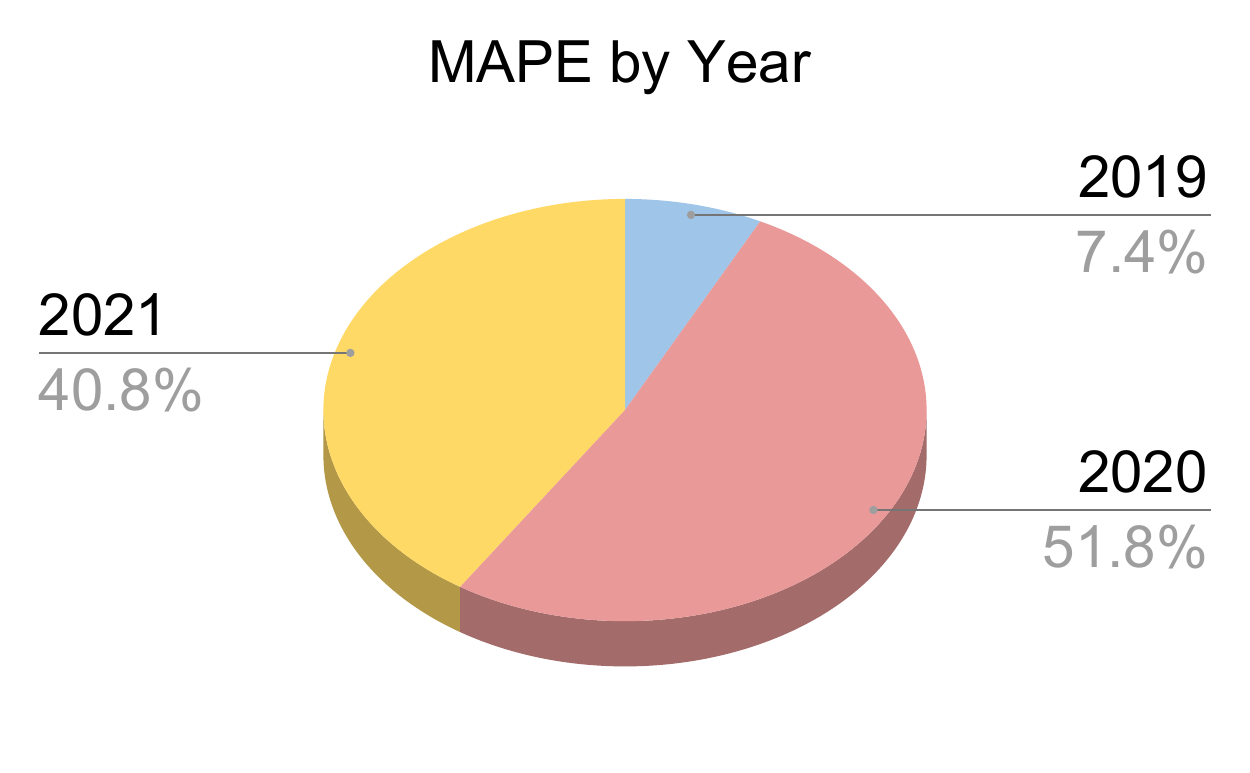}
        \caption*{(a) MAPE}
    \end{minipage}%
    \hspace{0.0125cm}
    \begin{minipage}{0.31\textwidth}
        \centering
        \includegraphics[width=\textwidth]{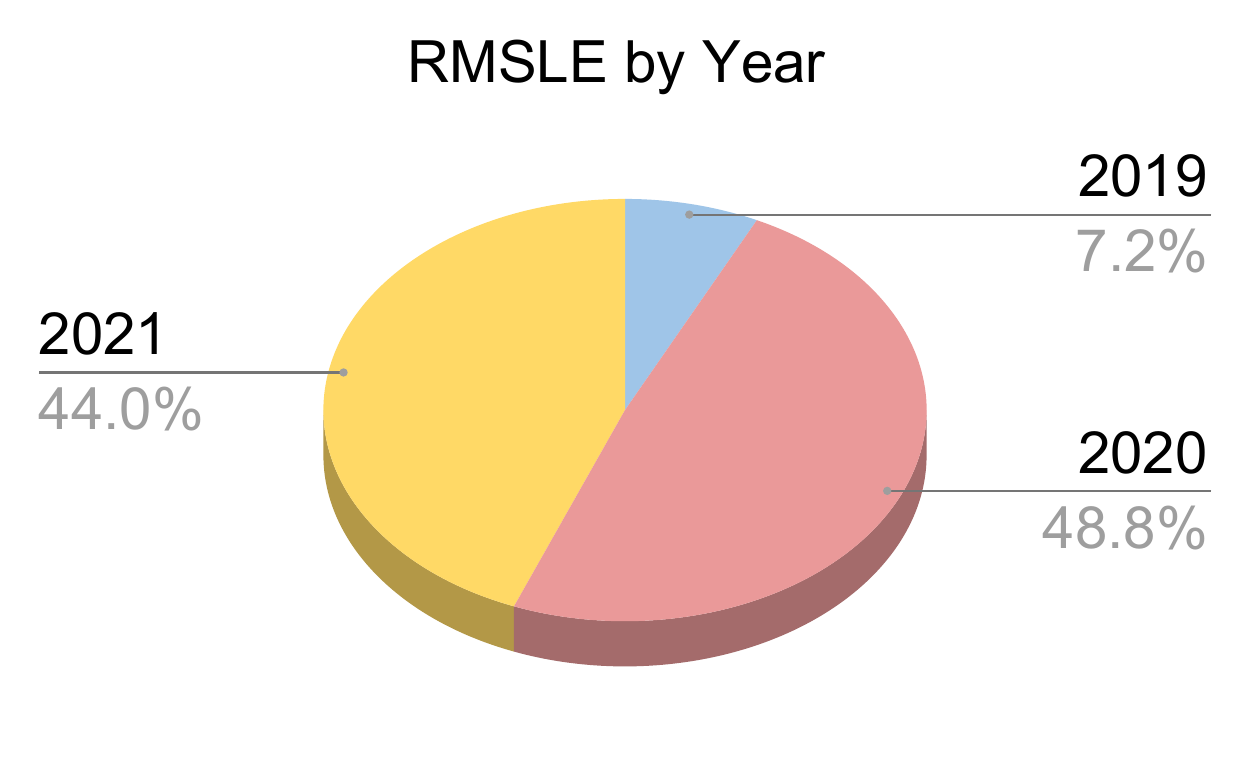}
        \caption*{(b) RMSLE}
    \end{minipage}%
    \hspace{0.0125cm}
    \begin{minipage}{0.31\textwidth}
        \centering
        \includegraphics[width=\textwidth]{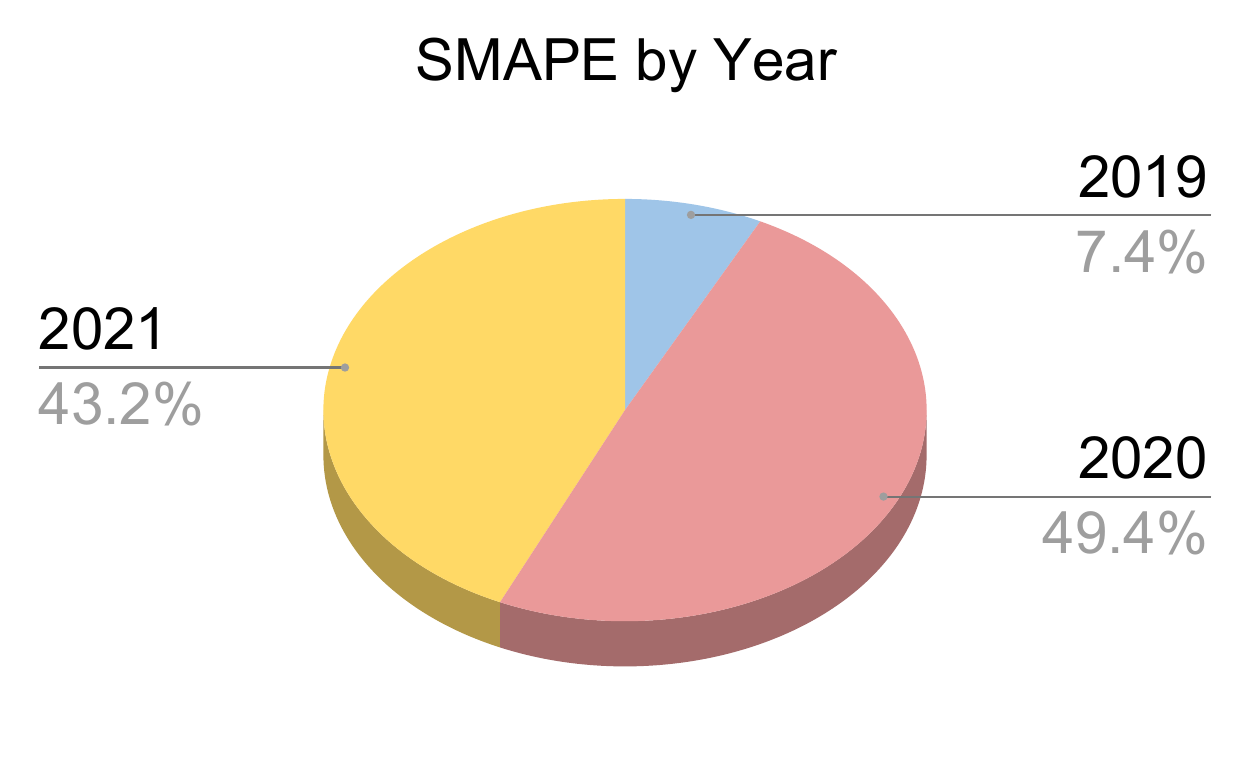}
        \caption*{(c) SMAPE}
    \end{minipage}
    \caption{ Comparison of metrics over the years for the Sentinel-2 dataset.}
  \label{fig: years_s2}
\end{figure}

The pie charts in Fig.~\ref{fig: years_s1} represent the distribution of MAPE, RMSLE, and SMAPE for Sentinel-1 from 2019 to 2021. In 2020, error rates surged, with MAPE at 48.30\%, RMSLE at 54.2\%, and SMAPE at 53.7\%, likely due to challenges in data collection and environmental factors impacting measurement accuracy.  
However, 2021 showed a slight recovery in error metrics, with MAPE decreasing to 41.0\% and RMSLE and SMAPE reducing to 37.2\% and 37.4\%, respectively, suggesting improvements in error correction and data processing.
In contrast, 2019 marked the best performance, with MAPE at 11.0\%, RMSLE at 8.6\%, and SMAPE at 8.9\%, likely due to more stable conditions or improved methodologies. Overall, the trends highlight the challenges faced in 2020 and the gradual recovery observed in 2021.
The pie charts in Fig.~\ref{fig: years_s2} illustrate the distribution of MAPE, RMSLE, and SMAPE for Sentinel-2 across the years 2019, 2020, and 2021. 
In 2020, error rates peaked significantly, with MAPE at 51.8\%, RMSLE at 48.8\%, and SMAPE at 49.4\%, indicating data accuracy challenges. However, 2021 showed improvement as MAPE decreased to 40.7\%, RMSLE to 44.0\%, and SMAPE to 43.2\%, likely due to refined data collection methodologies. In contrast, 2019 had the lowest error rates, with MAPE at 7.4\%, RMSLE at 7.2\%, and SMAPE at 7.4\%, setting a benchmark for accuracy. These fluctuations emphasize the need for continual optimization and effective error mitigation strategies in satellite monitoring systems such as Sentinel-2.
 
\begin{figure}[h!]
    \centering
    \begin{minipage}{0.31\textwidth}
        \centering
        \includegraphics[width=\textwidth]{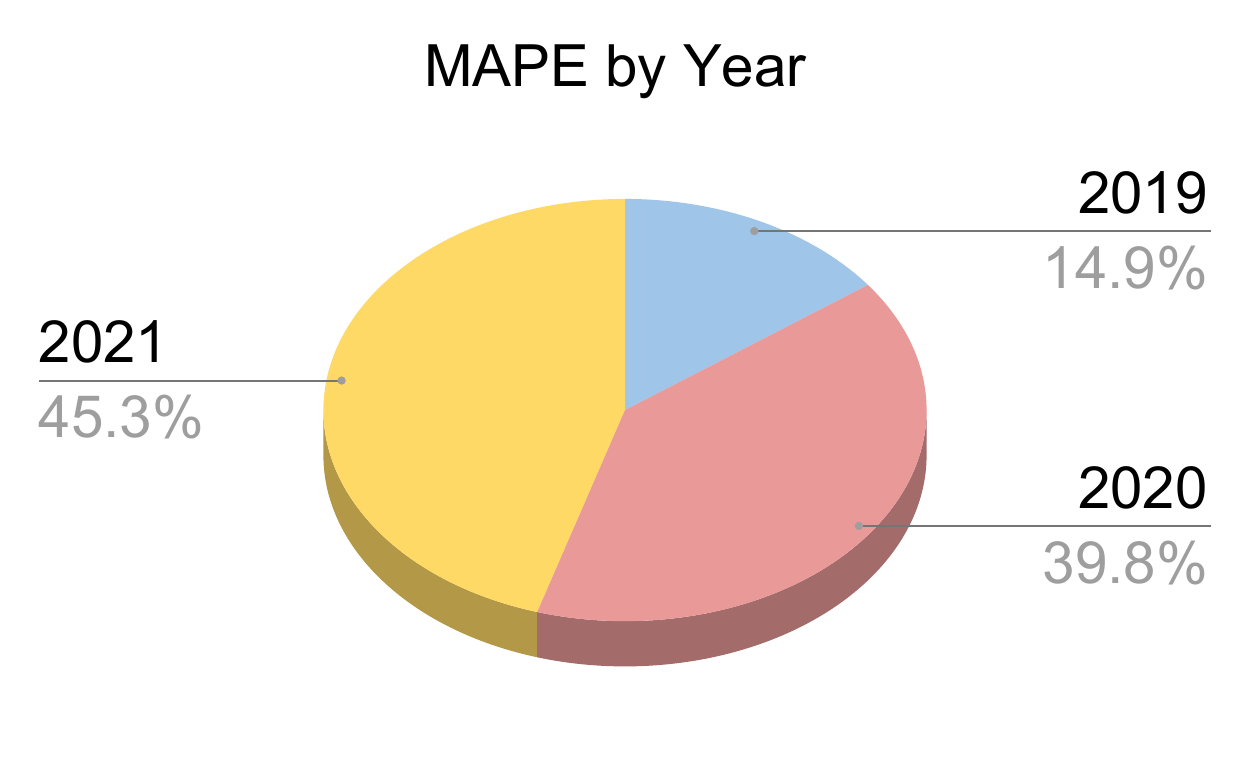}
        \caption*{(a) MAPE}
    \end{minipage}%
    \hspace{0.0125cm}
    \begin{minipage}{0.31\textwidth}
        \centering
        \includegraphics[width=\textwidth]{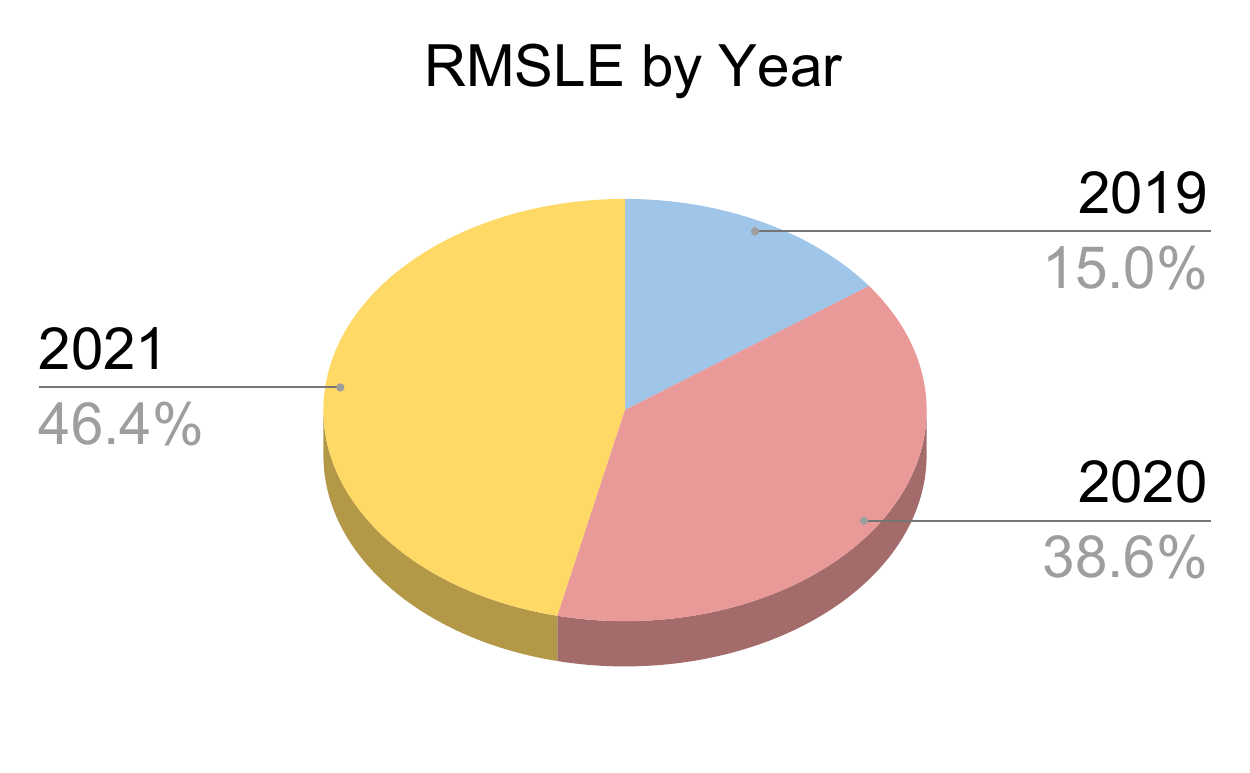}
        \caption*{(b) RMSLE}
    \end{minipage}%
    \hspace{0.0125cm}
    \begin{minipage}{0.31\textwidth}
        \centering
        \includegraphics[width=\textwidth]{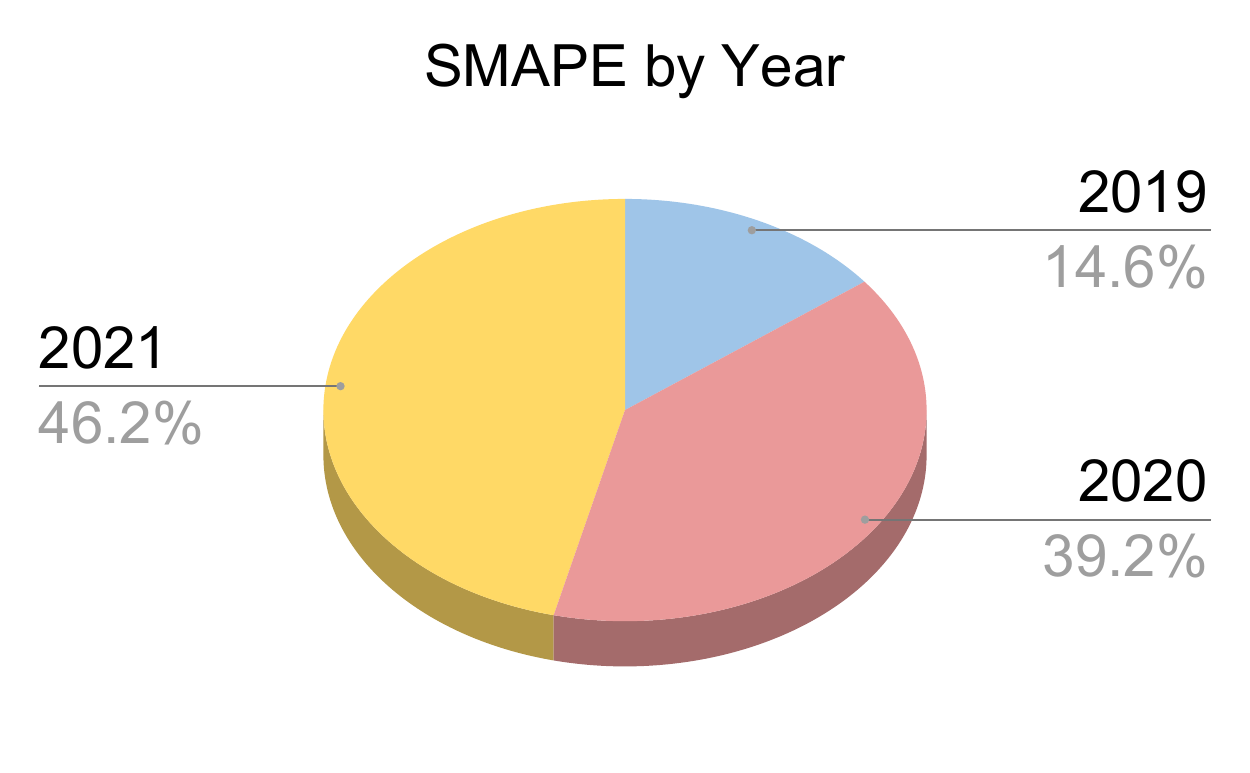}
        \caption*{(c) SMAPE}
    \end{minipage}
    \caption{ Comparison of metrics over the years for the Landsat-8 dataset.}
  \label{fig: years_l8}
\end{figure}

The pie charts in Fig.~\ref{fig: years_l8} illustrate the distribution of MAPE, RMSLE, and SMAPE for Landsat-8 from 2019 to 2021.
In 2021, the errors were relatively higher, with MAPE at 45.3\%, RMSLE at 46.4\%, and SMAPE at 46.2\%. The year 2020 shows a slightly lower distribution of errors, with MAPE at 39.8\%, RMSLE at 38.6\%, and SMAPE at 39.2\%. In contrast, 2019 exhibits the lowest error contributions, with MAPE at 14.9\%, RMSLE at 15.0\%, and SMAPE at 14.6\%, reflecting improved model accuracy in earlier years.

\subsection{Comparison of metrics over the seasons}
\begin{figure}[h!]
    \centering
    \begin{minipage}{0.32\textwidth}
        \centering
        \includegraphics[width=\textwidth]{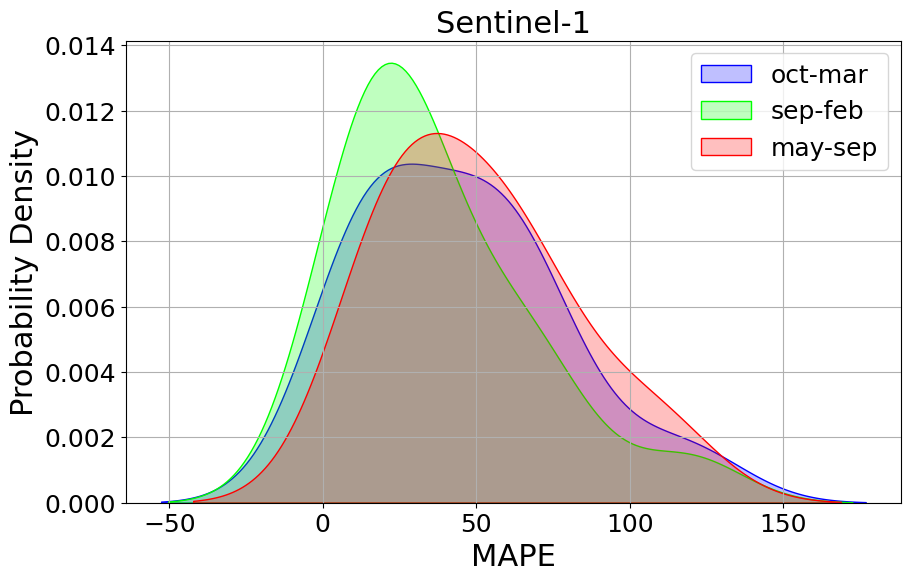}
    \end{minipage}%
    \hspace{0.2cm}
    \begin{minipage}{0.32\textwidth}
        \centering
        \includegraphics[width=\textwidth]{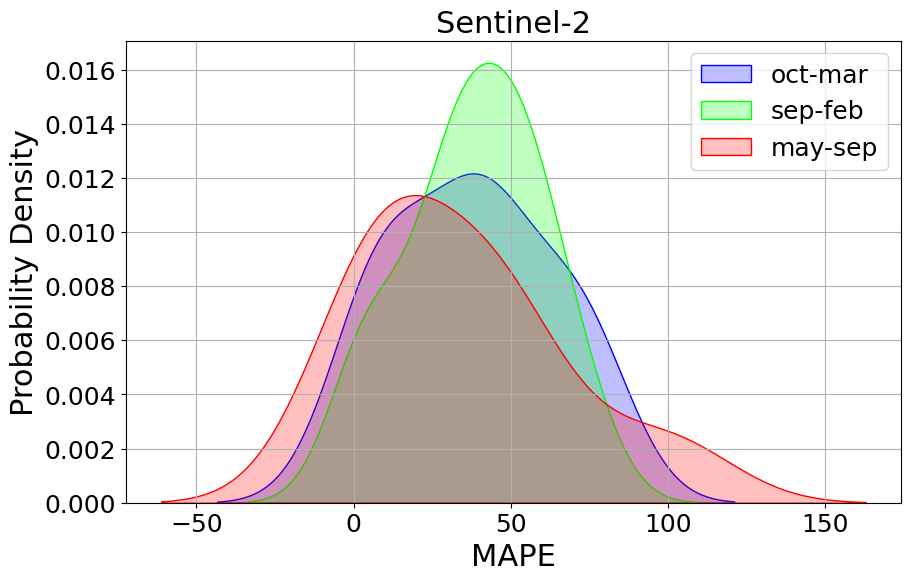}
    \end{minipage}%
    \hspace{0.2cm}
    \begin{minipage}{0.32\textwidth}
        \centering
        \includegraphics[width=\textwidth]{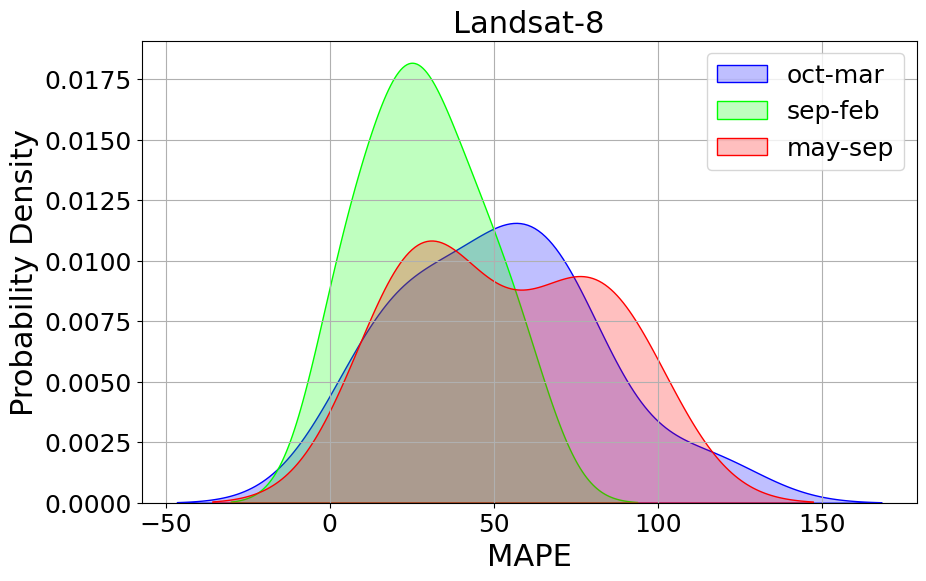}
    \end{minipage}
    \caption{ Comparison of MAPE over the seasons for the Sentinel-1, Sentinel-2, and Landsat-8 datasets.}
 \label{fig: season1}
\end{figure}

In Fig.~\ref{fig: season1}, density plots represent the MAPE performance of the Sentinel-1, Sentinel-2, and Landsat-8 datasets across different seasons, including Oct-Mar, Sep-Feb, and May-Sep. The first plot shows the Sentinel-1 model’s error distributions across different seasons. The MAPE density for May-Sep shows the greatest spread, indicating a higher likelihood of larger error values. The Sep-Feb period presents a more concentrated distribution, highlighting better alignment between predicted and actual values. In the second plot for Sentinel-2, the May-Sep season again shows the widest spread, with a higher probability density at larger error values, indicating less consistent model performance during this period. In contrast, the Sep-Feb season exhibits a narrower and sharper curve, reflecting improved prediction stability. The Oct-Mar season lies in between, with a broader spread than Sep-Feb but better overall consistency than May-Sep. 

The last plot in Fig.~\ref{fig: season1} corresponds to the Landsat-8 dataset. The MAPE density for May-Sep reflects a broader curve, signifying larger errors during this period. The Sep-Feb season remains the most consistent, with a narrower and more stable distribution, while Oct-Mar exhibits intermediate behavior.
 
\begin{figure}[h!]
    \centering
    \begin{minipage}{0.32\textwidth}
        \centering
        \includegraphics[width=\textwidth]{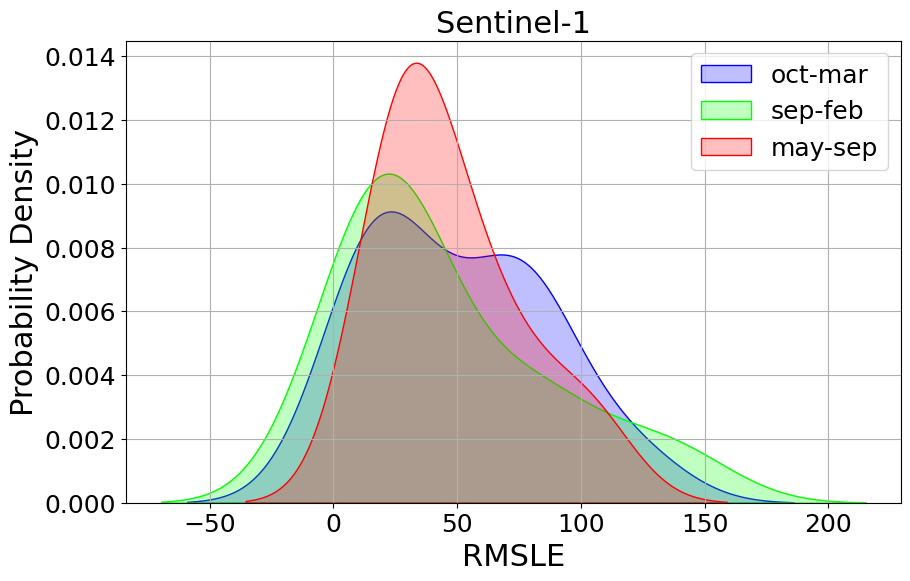}
    \end{minipage}%
    \hspace{0.2cm}
    \begin{minipage}{0.32\textwidth}
        \centering
        \includegraphics[width=\textwidth]{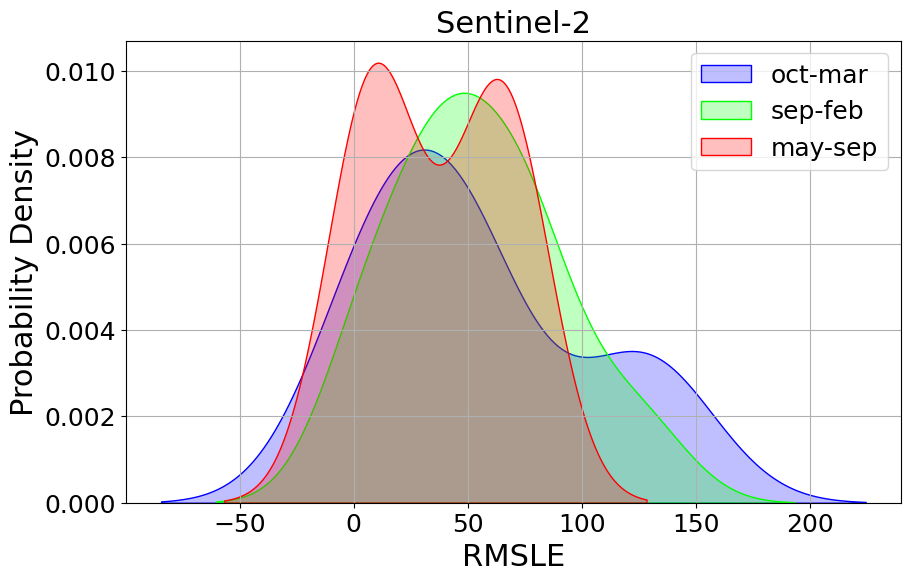}
    \end{minipage}%
    \hspace{0.2cm}
    \begin{minipage}{0.32\textwidth}
        \centering
        \includegraphics[width=\textwidth]{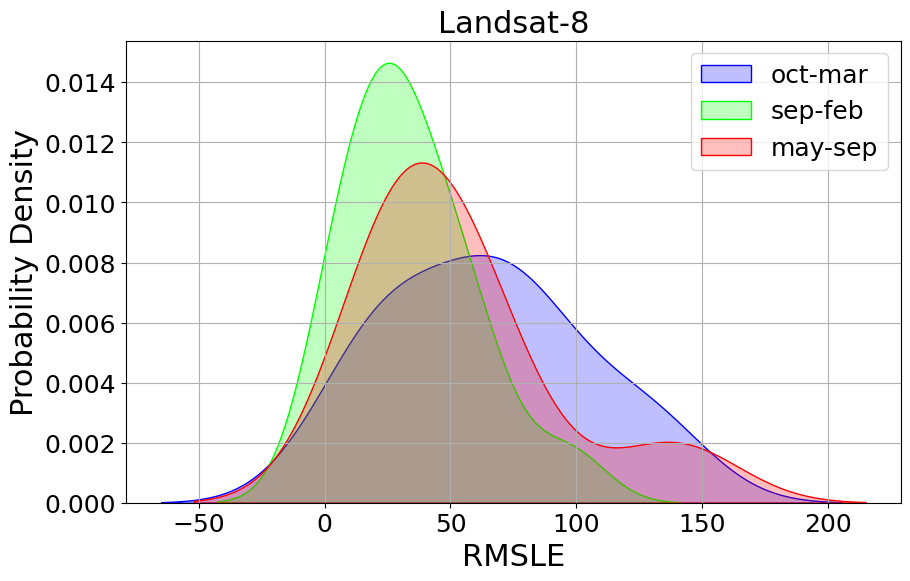}
    \end{minipage}
    \caption{ Comparison of RMSLE over the seasons for the Sentinel-1, Sentinel-2, and Landsat-8 datasets.}
  \label{fig: season2}
\end{figure}

In Fig.~\ref{fig: season2}, density plots visualize the performance of RMSLE for Sentinel-1, Sentinel-2, and Landsat-8 datasets across different seasons, including Oct-Mar, Sep-Feb, and May-Sep. The first density plot shows the Sentinel-1 dataset; the May-Sep season shows a left-skewed trend with higher deviations, while Sep-Feb achieves the most consistent alignment of predictions. The second density plot for the Sentinel-2 dataset May-Sep displays a similar left-skewed distribution with larger deviations between predictions and actual values, while Sep-Feb demonstrates a more concentrated density around lower error values. The third plot, representing Landsat-8, follows a similar pattern, with higher probability density for larger errors during May-Sep, emphasizing the seasonal challenges in maintaining prediction accuracy for this dataset. The Sep-Feb season consistently achieves the lowest RMSLE values, confirming better overall performance during this time.

\section{Conclusion}
\label{sec: conclude}
In this study, for identifying crop growth patterns over spatio-temporal variation with multi-spectral data, we propose the MTMS-YieldNet framework that incorporates spectral-special patterns with spatio-temporal dependencies.
Our framework performs two-step training, consisting of a pre-training step followed by final training. In the pre-training step, MTMS-YieldNet introduces a contrastive learning-based feature extraction method by leveraging two modules, namely the spatio-temporal dependency module, which captures temporal and spatial correlations, and the spectral-spatial attention module, which emphasizes the relevant features from multi-spectral data. Subsequently, feature selection improves performance by imitating natural balance processes and selecting key spectral and temporal features. The experimental results demonstrate that the proposed MTMS-YieldNet achieves good performance in crop yield prediction using multi-spectral imagery. The multi-spectral imagery helps to extract information on various interconnected factors that contribute to precise yield prediction, including vegetation indices, different properties of soil, and moisture content with spatio-temporal dependencies. In addition, the stacked attention mechanism refines spatial variations in soil and identifies temporal patterns in crop growth using multi-seasonal data, while multi-temporal analysis and soil-aware learning enhance crop yield estimation by incorporating dynamic growth trends and essential soil properties. Our proposed model enables accurate yield prediction efficiently, saving time and supporting farmers in their decision-making process.

\bibliographystyle{elsarticle-harv}
\bibliography{References}

\end{document}